\pdfoutput=1
\documentclass[11pt]{article}

\usepackage[preprint]{acl}

\usepackage{times}
\usepackage{latexsym}

\usepackage[T1]{fontenc}
\usepackage[utf8]{inputenc}

\usepackage{microtype}

\usepackage{inconsolata}

\usepackage{graphicx}

\usepackage{adjustbox}
\usepackage{listings}
\usepackage{pifont}
\usepackage{subcaption}
\usepackage{wrapfig}
\usepackage{xspace}
\usepackage[subtle]{savetrees}

\usepackage[utf8]{inputenc} % allow utf-8 input
\usepackage[T1]{fontenc}    % use 8-bit T1 fonts
\usepackage{url}            % simple URL typesetting
\usepackage{booktabs}       % professional-quality tables
\usepackage{amsfonts}       % blackboard math symbols
\usepackage{nicefrac}       % compact symbols for 1/2, etc.
\usepackage{microtype}      % microtypography
\usepackage{soul}
\usepackage{graphicx}
\usepackage{amsmath}
\usepackage{outlines}
\usepackage{xurl}

\usepackage{amsmath,amsfonts,bm}

\def\secref#1{Sec.~\ref{#1}}
\def\eqref#1{equation~\ref{#1}}
\def\1{\bm{1}}

\def\rmKV{{\mathbf{KV}}}

\def\rmM{{\mathbf{M}}}

\def\rmW{{\mathbf{W}}}

\DeclareMathAlphabet{\mathsfit}{\encodingdefault}{\sfdefault}{m}{sl}
\SetMathAlphabet{\mathsfit}{bold}{\encodingdefault}{\sfdefault}{bx}{n}

\usepackage{multirow}
\usepackage{paralist}

\usepackage{multicol}
\usepackage{diagbox}

\usepackage{here}

\usepackage{amsmath,amssymb,amsfonts,amsbsy,amsfonts,latexsym}
\usepackage{makecell}
\usepackage{xcolor}
\usepackage{colortbl}

\usepackage{tabularx,colortbl,xcolor}
\usepackage[normalem]{ulem}
\useunder{\uline}{\ul}{}

\usepackage{enumitem}

\usepackage{xparse}% http://ctan.org/pkg/xparse

\usepackage{longtable}

\NewDocumentCommand{\var}{O{s} m O{}}{%
  \ensuremath{#1_{#2}^{#3}}% add \vphantom{<bizarre sup>}
}
\usepackage{siunitx}

\newcommand{\commentout}[1]{}

\definecolor{light-gray}{gray}{0.80}

\newcommand\appref{Appendix~\ref}

\newcommand\fref{Fig.~\ref}
\newcommand\tref{Table~\ref}
\newcommand\sref{Sec.~\ref}

\def\x{{\bf x}}
\def\y{{\bf y}}

\usepackage{amsthm}

\usepackage{amsthm}

\newtheorem{mytheorem}{Theorem}
\newtheorem{assumption}[mytheorem]{Assumption}
\newtheorem{lemma}[mytheorem]{Lemma}
\newtheorem{remk}[mytheorem]{Remark}

\newcommand{\cmark}{{\ding{51}}}
\newcommand{\xmark}{{\ding{55}}}

\newcommand{\OURS}{SwiftKV\xspace}
\newcommand{\llm}{LLM\xspace}

\newcommand{\kv}{KV cache\xspace}

\newcommand{\llamasmall}{Llama-3.1-8B-Instruct\xspace}
\newcommand{\llamalarge}{Llama-3.1-70B-Instruct\xspace}

\newcommand{\singlekv}{SwiftKV\xspace}
\newcommand{\acrosskv}{AcrossKV\xspace}

\newcommand{\gqa}{GQA\xspace}
\newcommand{\gsm}{GSM-8K\xspace}

\title{\OURS: Fast Prefill-Optimized Inference with Knowledge-Preserving Model Transformation}

\author{
Aurick Qiao \qquad Zhewei Yao \qquad Samyam Rajbhandari \qquad Yuxiong He \\
\\
Snowflake AI Research \\ San Mateo, CA, United States \\
\small{
\textbf{Correspondence:} \href{mailto:aurick.qiao@snowflake.com}{aurick.qiao@snowflake.com}
}
}

\begin{document}
\maketitle

\begin{abstract}
LLM inference for enterprise applications, such as summarization, RAG, and code-generation, typically observe much longer prompt than generations, leading to high prefill cost and response latency.
We present \OURS, a novel model transformation and distillation procedure targeted at reducing the \emph{prefill compute} (in FLOPs) of prompt tokens while preserving high generation quality.
First, \OURS prefills later layers' KV cache using an earlier layer's output, allowing prompt tokens to skip those later layers.
Second, \OURS employs a lightweight knowledge-preserving distillation procedure that can adapt existing LLMs with minimal accuracy impact.
Third, \OURS can naturally incorporate \kv compression to improve inference performance in low-memory scenarios.
Our comprehensive experiments show that \OURS can effectively reduce prefill computation by 25--50\% across several LLM families while incurring minimum quality degradation.
In the end-to-end inference serving, \OURS realizes up to \(2\times\) higher aggregate throughput and 60\% lower time per output token. 
It can achieve a staggering 560 TFlops/GPU of normalized inference throughput, which translates to 16K tokens/s for Llama-3.1-70B. 
\OURS is open-sourced at \url{https://github.com/snowflakedb/arctictraining}.
\end{abstract}
\section{Introduction}
\label{sec:intro}

Large Language Models (LLMs) are now an integral enabler of enterprise applications and offerings, including code and data co-pilots~\citep{chen2021evaluatinglargelanguagemodels,10.5555/3666122.3667699}, retrieval augmented generation (RAG)~\citep{10.5555/3495724.3496517,lin2024radit}, summarization~\citep{pu2023summarizationalmostdead,zhang-etal-2024-benchmarking}, and agentic workflows~\citep{wang2024mixtureofagentsenhanceslargelanguage,schick2023toolformerlanguagemodelsteach}.  However, the cost and speed of inference determine their practicality, and improving the throughput and latency of LLM inference has become increasingly important.

While prior works, such as model pruning~\cite{ma2023llmpruner,sreenivas2024llmpruningdistillationpractice}, KV cache compression~\cite{hooper2024kvquant10millioncontext,shazeer2019fasttransformerdecodingwritehead,ainslie2023gqatraininggeneralizedmultiquery,chang2024palucompressingkvcachelowrank}, and sparse attention~\cite{zhao2024alisaacceleratinglargelanguage,jiang2024minference10acceleratingprefilling}, have been developed to accelerate LLM inference, they typically significantly degrade the model quality or work best in niche scenarios, such as low-memory environments or extremely long contexts requests (e.g. >100K tokens). On the other hand, production deployments are often compute-bound rather than memory-bound, and such long-context requests are rare amongst diverse enterprise use cases (e.g. those observed at Snowflake).

%The cost and speed of LLM inference is determined by its throughput and latency. High overall throughput i.e. aggregate tokens processed and generated per second per GPU results in a lower cost, while low latency per generated token and the time to generate the first results in faster inference speed. 

In this paper, we take a different approach to improving LLM inference based on the key observation that typical enterprise workloads process more input tokens than output tokens. For example, tasks like code completion, text-to-SQL, summarization, and RAG each submit long prompts but produce fewer output tokens (a 10:1 ratio with average prompt length between 500 and 1000 is observed in our production). In these scenarios, inference throughput and latency are often dominated by the cost of prompt processing (i.e. prefill), and reducing this cost is key to improving their performance.

Based on this observation, we designed \emph{\OURS}, which improves throughput and latency by reducing the prefill computation for prompt tokens. \OURS (\fref{fig:main-figure}) consists of three key components:

\begin{figure*}
\centering
\includegraphics[width=0.9\textwidth,trim={0 150 80 0},clip]{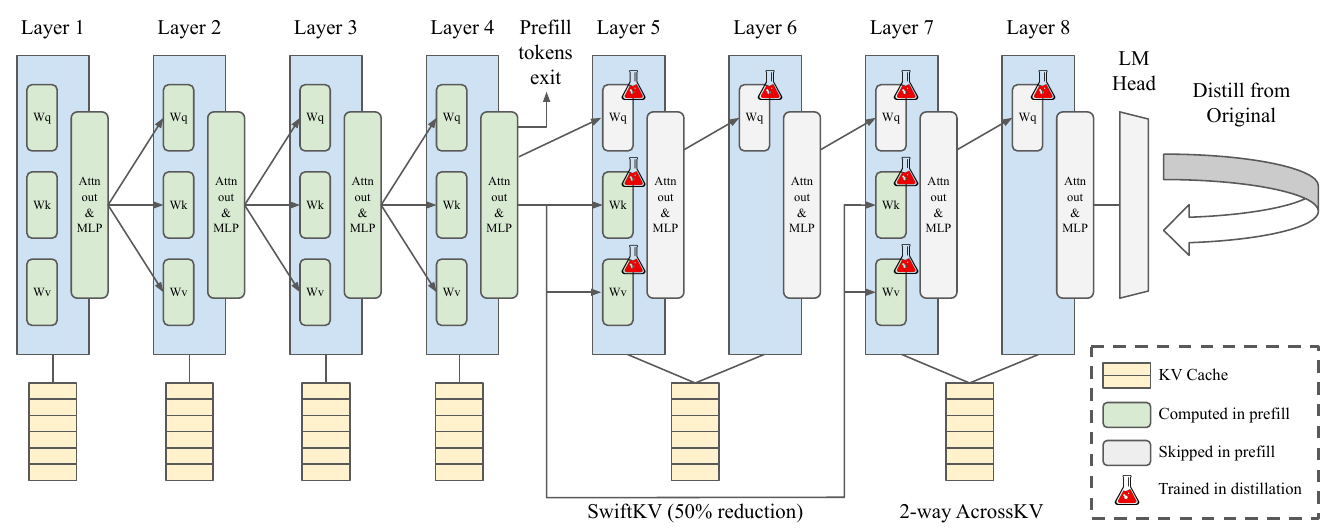}
\caption{
Illustration of \OURS 50\% prefill reduction and 2-way \acrosskv. After distillation, the KV cache of layers 5--8 can all be populated using the hidden state outputs of layer 4. For prefill tokens, the query, attention, and MLP operations of layers 5--8 may be skipped, while decode tokens complete all layers. Existing models may be efficiently adapted for \OURS by distilling from the original model using a small dataset. Model knowledge is preserved by keeping the trainable parameters limited to the Q, K, and V projections of the layers affected by \singlekv.
}
\label{fig:main-figure}
\end{figure*}

\paragraph{Model transformation.}
\singlekv rewires an existing LLM so that the prefill stage during inference can skip a number of later transformer layers, and their KV cache are computed by the last unskipped layer.
This is motivated by the observation that the hidden states of later layers do not change significantly (see \sref{sec:single_layer_kv} and \citep{liu2024foldgptsimpleeffectivelarge}).
With \singlekv, prefill compute is reduced by approximately the number of layers skipped. 

Optionally, for low-memory scenarios, we show that the \OURS model transformation can naturally incorporate \kv memory reductions by also merging the \kv of consecutive skipped layers, which we call \acrosskv.

\paragraph{Knowledge Recovery.} After the \singlekv transformation is applied to the LLM, its prediction quality is recovered via distillation from the original model. A very lightweight distillation is sufficient, with <10\% the model weights (Q, K, and V projections of the skipped layers) trained on <1B tokens, which takes less than 3 hours on 8 H100 GPUs for Llama-3.1-8B-Instruct.
In contrast, recent prune-and-distill techniques train the entire pruned model on 10--100B tokens~\citep{tang2025darwinlmevolutionarystructuredpruning, sreenivas2024llmpruningdistillationpractice}.

We show that \OURS is effective on diverse architectures, including small models (Llama-3.2-3B-Instruct), large models (Llama-3.1-405B-Instruct), mixture-of-experts and latent attention (Deepseek-V2-Lite-Chat). Remarkably, we found that it is possible for \OURS to skip 25--50\% of the layers for prompt tokens without significantly impacting the model quality across these scenarios.

 %It is possible to directly use the KV caches from the $n^{th}$ layer to reduce both pre-fill computation and KV cache footprint, however, we find that without leveraging all of the KV projections, the model quickly looses its knowledge from discarding parameters (see Fig. XYZ). 

%Instead, we develop flexible KV cross layer sharing (Sec XYZ) and combine it with low-precision quantization to achieve up to Mx of KV cache reduction in conjunction with the reduction in pre-fill computation (see Fig. XYZ). 

%ii) Given the first observation, reducing the computation needed to process the context/pre-fill can significantly reduce the total computation for inference. 

% using parameter-preserving-compute-lowering model transformation. \OURS is a model distillation approach, where the student has the same set of parameters as the teacher, re-wired with the goal of reducing computation. As such, the knowledge of the teacher model is preserved in the student model parameters, and unlike many model distillation approach where the distillation is used for knowledge or skill transfer, \OURS uses the distillation to learn how to extract its existing knowledge and skills from the re-configured architecture.

\paragraph{Optimized Inference.} To realize \OURS into end-to-end throughput and latency improvements, we implemented it in vLLM~\citep{10.1145/3600006.3613165} and SGLang~\cite{zheng2024sglang}.
%Our implementations include additional optimizations, including fusing all KV-projections beyond layer $l$ into a single GEMM operation, and integrated memory management needed to lower the KV cache memory footprint achievable via \acrosskv.
\OURS increases the throughput of enterprise workloads by up to \( 2\times \), while reducing time-to-first-token (TTFT) and time-per-output-token (TPOT) by up to 50\% and 60\%, respectively. In fact, for Llama-3.1-70B-Instruct, \OURS achieves a normalized throughput of 560 TFLOPS/GPU\footnote{Normalized throughput and MFU is based on number of floating point operations in the baseline model.} at an unprecedented 56.6\% MFU utilization for inference (\secref{sec:inference-speedup}). \OURS incurs minimal quality degradations (<1--2\%) averaged across a wide range of tasks (\secref{subsec:model-quality-vs-compression}), including ARC-Challenge~\citep{Clark2018ThinkYH}, Winogrande~\citep{sakaguchi2019winogrande}, HellaSwag~\citep{zellers2019hellaswag}, TruthfulQA~\citep{lin-etal-2022-truthfulqa}, MMLU~\citep{hendryckstest2021}, and GSM8K~\citep{cobbe2021training}.

%In addition to these main results, in \sref{sec:discussion} we discuss the impact of distillation, datasets, choice of trainable parameters for training \OURS. We also present our analysis of the hidden state similarities, and how \acrosskv can be extended and combined with other KV cache compression works. Additionally, we also discuss how \singlekv can enable compute savings not just during pre-fill but also during decoding phase.

We open-sourced the training code for \OURS at \url{https://github.com/snowflakedb/arctictraining} and and inference code at \url{https://github.com/snowflakedb/arcticinference}, as well as several \OURS models that can be used directly by the community at \url{https://huggingface.co/collections/Snowflake/swiftkv-models-674f7d7474eb789e185d31cb}.

%To offer better insights into the impact of distillation and datasets, similarity of hidden states across layers, how \OURS can be combined with other KV cache compression works, and enhanced with early exit for decoding, we share ablations results and discussions in Sec.~\ref{ablations-and-discussion}.  

% \zhewei{I have a overall different favor to write the intro and I put outline here}
% 1. KV Cache is the main bottleneck for fast and efficient inference.

% 2. From fast inference perspective, system work + some weight quantization work has been proposed. However, no one tackle the KV cache projection from early layers first

% 3. From kv reduction for larger batch size, quantization work and MiniCache are proposed. People also do GQA etc but we propose a new setting and highlight the shortcoming of Minicache.

% Taking all these into consider. We propose \OURS, which inlcudes the following contributions:

% 1. \singlekv

% 2. \acrosskv

% 3. knowledge-preserved distillatoin

% 4. Full vLLM implementation and real speedup 

% 5. Highlight some main results and ablation results

\section{Related Works}
\label{sec:related_work}

%\textbf{Hardware and system optimizations.} Lower-precision quantization like FP8~\citep{10.5555/3600270.3601335} can enable the use of tensor-cores to accelerate inference~\citep{luo2024benchmarkingdissectingnvidiahopper}. System methods like PagedAttention~\citep{10.1145/3600006.3613165}, Tensor-Parallelism~\citep{shoeybi2020megatronlmtrainingmultibillionparameter}, Split-Fuse~\citep{holmes2024deepspeedfastgenhighthroughputtextgeneration,298679}, FlashAttention~\citep{10.5555/3600270.3601459}, and their implementations in  TensorRT~\citep{TensorRT}, FasterTransformer~\citep{FasterTransformer}, vLLM~\citep{10.1145/3600006.3613165}, and DeepSpeed-Inference~\citep{aminabadi2022deepspeedinferenceenablingefficient} enable better parallelization, batching, and scheduling to eliminate system overheads and achieve better hardware utilization without impacting model quality. In contrast, \OURS is a model architecture optimization and is complementary to these works.

\paragraph{Model pruning and layer skipping.}
Prior works have explored reducing the size and compute footprint of LLMs by pruning their weights, followed by post-training on 10--100B tokens to recover accuracy~\citep{tang2025darwinlmevolutionarystructuredpruning, sreenivas2024llmpruningdistillationpractice, xia2024shearedllamaacceleratinglanguage}. Compared to these methods, \OURS is focused on reducing prefill compute using a much lighter-weight distillation (<1B tokens). Other works explored adaptively skipping layers without pruning weights, and using little to no post-training~\cite{ma2023llmpruner, jaiswal2024ffnskipllmhiddengemautoregressive, men2024shortgptlayerslargelanguage, yang2024lacolargelanguagemodel, ashkboos2024slicegptcompresslargelanguage}. These works reduce compute for prefill and decode tokens alike, and typically can skip up to 25\% of the model without significant accuracy degradations. \OURS reduces prefill compute, and can skip 25--50\% of the model without significant accuracy degradations.

\paragraph{\kv compression.}
%A wide range of techniques have been developed to reduce the memory need of the \kv.
Quantization techniques like FP8/FP4 can reduce the memory for both \kv and parameters~\citep{hooper2024kvquant10millioncontext}. Attention optimizations like MQA~\citep{shazeer2019fasttransformerdecodingwritehead}, GQA~\citep{ainslie2023gqatraininggeneralizedmultiquery}, low-rank attention~\citep{chang2024palucompressingkvcachelowrank} also reduce the \kv.
These approaches are complementary to \OURS, which we demonstrate in \sref{subsec:model-quality-vs-compression} and \sref{subsec:kv-quantization}.
%Like \acrosskv, MiniCache~\citep{liu2024minicachekvcachecompression} also considers merging the \kv of consecutive layers. However, \acrosskv enables consolidating more than just two layers, allowing for higher level of compression, and does not require any token retention strategy where distinct KV caches are stored for special tokens, allowing for simpler implementation.
Furthermore, while many of these approaches only focus on reducing the memory, \OURS reduces both the prefill compute and memory (via \acrosskv). As we show in \sref{sec:compute-vs-memory}, compute reduction is crucial for accelerating LLM inference in compute-bound scenarios with sufficient memory, which is common in production with modern GPUs (e.g., A100, H100).

\paragraph{Sparse attention.} Systems such as ALISA~\citep{zhao2024alisaacceleratinglargelanguage} and MInference~\citep{jiang2024minference10acceleratingprefilling} leverage naturally-occurring sparsity patterns in transformer models to reduce the computation of the quadratic attention operation. Sparse attention can be particularly effective for very long sequence lengths (e.g. 100K--1M tokens) when attention is the dominant operation.
In comparison, \OURS reduces prefill computation by skipping not just the attention operation, but also the query/output projections and MLP of certain layers. This means that \OURS can be more suited for inputs with moderate lengths (e.g. <100K) when MLP is the dominant operation. Additionally, \OURS either runs or skips attention operations in their entirety, which makes it orthogonal to existing sparse attention methods.

\section{\OURS: Design and Implementation}
\label{sec:methodology}

\begin{figure*}
\centering
\includegraphics[width=0.35\linewidth]{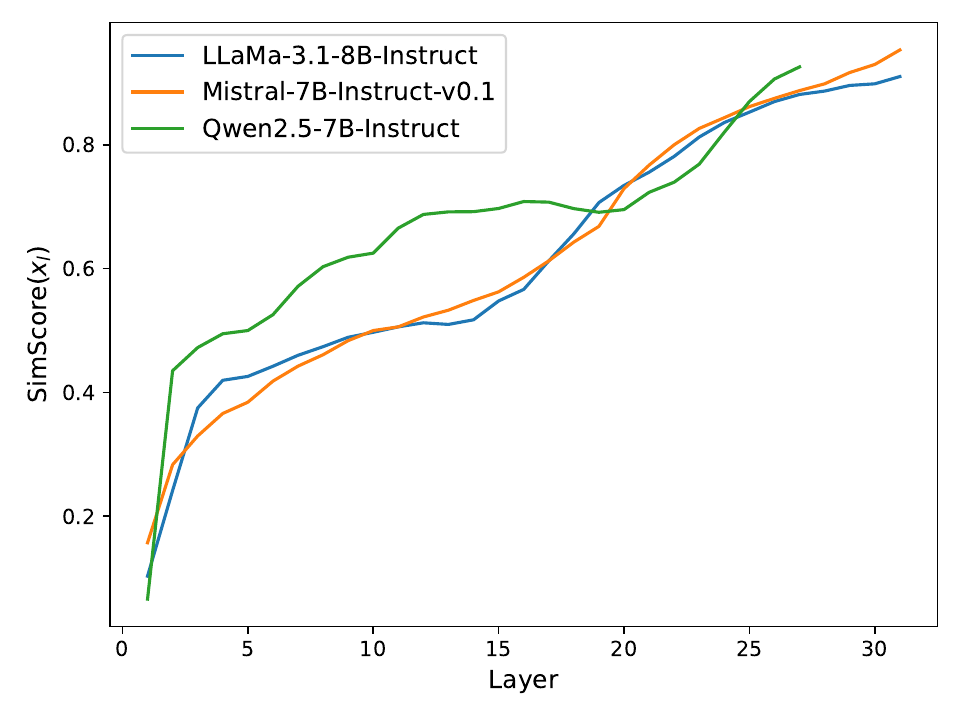}
\includegraphics[width=0.35\linewidth]{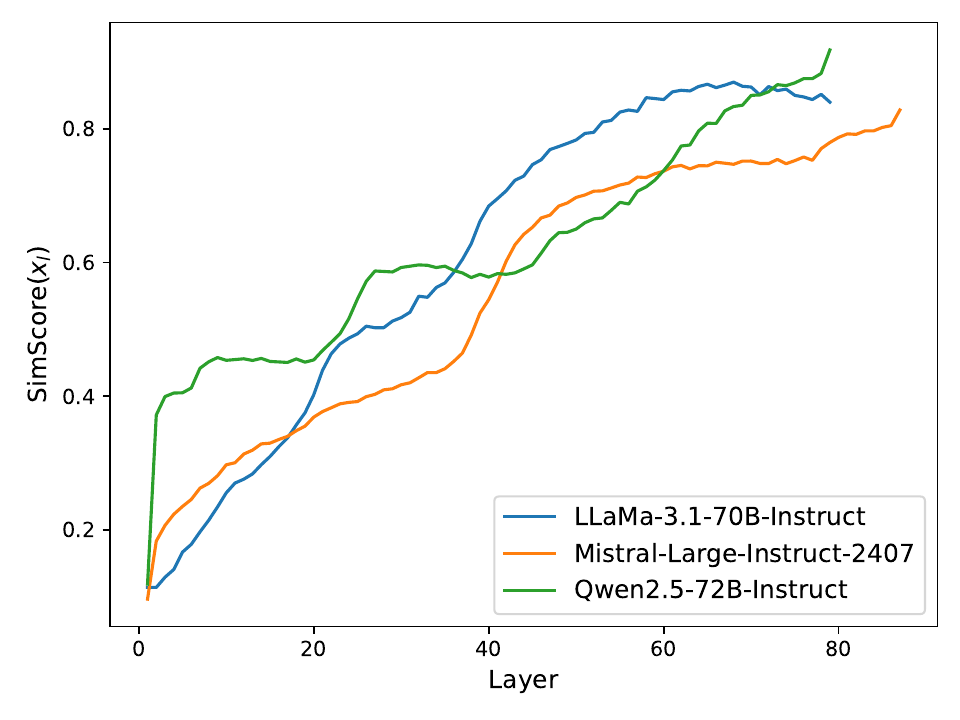}
\includegraphics[width=0.27\linewidth,trim={25 520 330 30},clip]{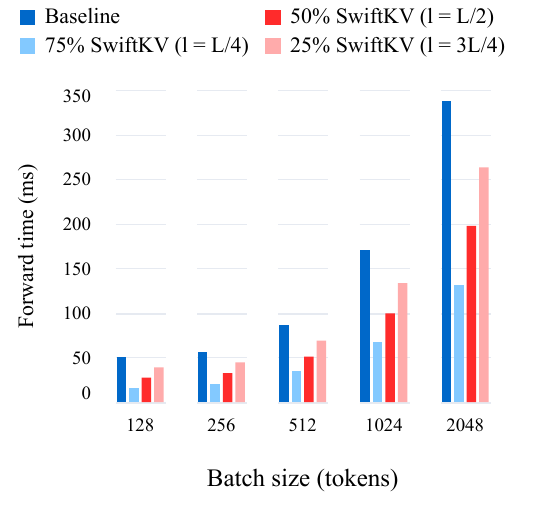}
\caption{
LEFT: input similarity of smaller models. MID: input similarity of larger models. RIGHT: time per forward pass of Llama-3.1-8B-Instruct. \singlekv reduces the forward pass processing time across a range of batch sizes.
%To estimate $\mathrm{SimScore}(\x_l)$, we use 50 examples from \texttt{HuggingFaceH4/ultrachat\_200k}.
}
\label{fig:motivation}
\end{figure*}

%{\color{blue}[YH:  This is probably the most important feedback I view - we can do better to incorporate the background and context of transformer, KV cache, etc. to make the paper more self-contained and easier to read / appreciate.  For example, it is helpful to introduce attention computation, what kv cache stores and why it is impportant.  It will also be helpful to show kv projection vs the rest of the computation which can be utilized to concretely point out the computation savings of our approach.  It will also be helpful to show at GQA, why kv projection is only a small portion of the total computation (comparing with the original).  How to introduce these context properly needs some thoughts: some may appear in background (we may have a mixed background + related work section), and some can show up at the corresponding design section. } --> {\color{red} SR: Addressed with a paragraph discuss this in the intro in SingleInputKV subsection}

\subsection{Preliminaries}

In transformers~\citep{10.5555/3295222.3295349}, attention enables each token to focus on other tokens by comparing \emph{queries} (\(Q\)) with \emph{keys} (\(K\)), using \emph{values} (\(V\)) to compute the final representation. For a sequence of input tokens \(x^{(1)}, \dots, x^{(n)}\), the projections are: $Q = X W_Q$, $K = X W_K$, $V = X W_V$, where \(X \in \mathbb{R}^{n \times d}\) are the input embeddings, and \(W_Q \in \mathbb{R}^{d \times d_k}\) and \( W_K, W_V \in \mathbb{R}^{d \times d_g}\) are trained model parameters with \( d_g | d_k \). Hereafter, we may also refer to \( W_K \) and \( W_V \) as a single matrix \( W_{KV}\in \mathbb{R}^{d \times 2 d_k} \).

During the \emph{prefill phase} of inference, the model processes the entire input sequence, computing \(K\) and \(V\) for all tokens in parallel (or in chunks in the case of Split-Fuse~\citep{holmes2024deepspeedfastgenhighthroughputtextgeneration,298679}). This typically occurs when the model handles an initial prompt or context.

During the \emph{decoding phase} of inference, new tokens are generated one at a time. When predicting the next token, only the query (\(Q^{(t+1)}\)) for the new token needs to be computed, while the model attends to the keys and values (\(K^{(1)}, \dots, K^{(t)}\), \(V^{(1)}, \dots, V^{(t)}\)) of all previous tokens.

In the decoding phase, \emph{KV caching} is employed. After processing each token \(t\), the newly computed \(K^{(t)}\) and \(V^{(t)}\) are stored in a cache. For the next token \(t+1\), only the new query \(Q^{(t+1)}\), key \(K^{(t+1)}\), and value \(V^{(t+1)}\) are computed. The attention computation will then utilize the cached \(K\) and \(V\) from all prior tokens, allowing for reduced computational overhead during inference.

\subsection{\singlekv: Project \kv from one layer}
\label{sec:single_layer_kv}

Assume the input of $l$-th layer is $\x_l$, and its $i$-th token is $\x_{l}^{(i)}$.
A key property of LLMs is that $\x_l$ becomes more similar as the depth grows~\citep{liu2024foldgptsimpleeffectivelarge,gromov2024unreasonableineffectivenessdeeperlayers}.

To illustreate, we compute the average input similarity between $l$-th layer's input and all remaining layers' input, i.e.,
\begin{equation}
    \mathrm{SimScore}(\x_l) = \frac{\sum_{j=l+1}^L\mathrm{Similarity}(\x_l, \x_j)}{L-l},
\end{equation}
where $L$ is the number of layers and $\mathrm{Similarity(\x_l,\x_j)}$ is the average cosine similarity between all $\x_{l}^{(i)}$ and $\x_{j}^{(i)}$.

The results of several models are shown in \fref{fig:motivation}.
Deeper layers have higher $\mathrm{SimScore}(\x_l)$, and at around half of the depth, the average similarity of $\x_l$ with $\x_{>l}$ is above \( 0.5 \) for all models, which shows that the difference of input hidden states are small in deeper layers.

Based on this observation, the first key component of \OURS is to use $l$-th layer's output $\x_{l+1}$ to compute the \kv for all remaining layers. More specifically, \OURS retains the standard transformer architecture up to and including the $l$-th layer, but the \kv for all remaining layers are computed immediately using $\x_{l+1}$, i.e.
\begin{equation}
\label{eq:singlekv}
    \rmKV_{j} = \rmW_{KV}^j\x_{l+1},~~~~\text{for all $j > l$,}
\end{equation}
%{\color{blue} [YH:for the last layer of computing KV cache properly, let's fix the notation.  We are using $l$ here and using $n$ earlier.]}-->{\color{red} SR:Done}
where $\rmKV_j$ is the \kv for $j$-th layer and $\rmW_{KV}^j$ is its KV projection weight matrix.
%As such, for all remaining layers ($> l$), there is no KV projection inside layers anymore. 
%In practice, we can concatenate all $\rmW_{KV}^j$ into one large weight matrix $\rmW_{KV}^{l+1, \ldots, L}$ and compute \(\rmKV_{l+1}, \ldots, \rmKV_{L}\) with a single efficient operation (\sref{}), and we refer this as \singlekv.
%{\color{blue} [YH: Can we align the representation at equation 2 with the notations at preliminaries? 
 %For example, preliminaries use $W_K$ and $W_V$ but we are using $W_{KV}$ here.]} \aurick{updated preliminaries with this alternative notation}
%--> {\color{red} Done in the last paragraph by quantifying the reduction and pointing to the detailed paragraph in the intro}

%\zhewei{We need a figure to show the key design of our method.}
\paragraph{Prefill Compute Reduction.} \singlekv enables significant reduction in prefill compute during \llm inference. Originally, all input tokens must be processed by all transformer layers.
With \singlekv, input tokens\footnote{The very last input token still needs to compute all layers to generate the first output token.} only need to compute \( \rmW_{KV}^j\x_{l+1} \) for layers $j > l$ to generate layer \( j \)'s \kv, and all other operations (i.e., QO projections, Attention, and MLP) of layers \( j > l \) can be skipped entirely. When prefill computation dominates generated token computation, this reduces the total inference computation to approximately $l/L$. \fref{fig:main-figure} illustrates the operations skipped by \singlekv, and \tref{tab:compute_breakdown} shows a more detailed example compute breakdown for Llama-3.1-70B-Instruct. We note that decoding tokens still propagate through all layers, so additional decoding heads are not necessary for \OURS.

\begin{table}%{r}{\linewidth}
%\centering
\caption{
Breakdown of transformer operations for Llama-3.1-70B with \OURS (in GFlops per prefill token).
}
\label{tab:compute_breakdown}
\begin{adjustbox}{width=\linewidth}
\footnotesize
\begin{tabular}{lccccccc}
\toprule
Model & Vocab & K,V & Q,O & MLP & Attn. & Total & Rel. \\
\midrule
\vspace{1mm}Baseline & 4.3 & 2.6 & 22 & 113 & 160 & 302 & 100\% \\
\vspace{1mm}25\% \singlekv & 4.3 & 2.6 & 16 & 85 & 120 & 228 & 75.5\% \\
\vspace{1mm}50\% \singlekv & 4.3 & 2.6 & 11 & 56 & 80 & 154 & 51.0\% \\
50\% \singlekv & \multirow{2}{*}{4.3} & \multirow{2}{*}{1.7} & \multirow{2}{*}{11} & \multirow{2}{*}{56} & \multirow{2}{*}{80} & \multirow{2}{*}{153} & \multirow{2}{*}{50.7\%} \\
+ \(4\times\) AcrossKV & & & & & & & \\
\bottomrule
\end{tabular}
\end{adjustbox}
\end{table}

\subsection{\acrosskv: Share \kv between layers}

%As shown in~\cite{liu2024minicache}, majority of the \kv can be merged for two adjacent layers.
GQA~\citep{ainslie-etal-2023-gqa}, one of the most widely adopted \kv compression methods, showed that the \kv can be shared across attention heads within the same transformer layer. Later, \cite{liu2024minicache} showed that the \kv can be merged for certain pairs of adjacent layers.
Although \OURS's main focus is on compute reduction rather than memory reduction, we show that KV cache compression can readily be incorporated with \OURS.
To do this, \OURS is supplemented by \acrosskv, which employs cross-layer \kv sharing to the skipped layers. 

Particularly, instead of computing \kv for all of the skipped layers as shown in~\eqref{eq:singlekv}, \acrosskv selects one layer to compute the \kv for several consecutive layers and share it within the small group (\fref{fig:main-figure}). \acrosskv can combine more than two layers' KV caches into a single one, which offers higher potential compression ratios than prior works~\cite{liu2024minicache} that employ cross-layer \kv merging, while simplifying its implementation.
%(See Sec. ~\ref{sec:related_work} for more detailed comparison with \cite{liu2024minicache}).
%Though our idea is similar to~\citep{liu2024minicache} but there are a few fundamental differences, including
%\begin{itemize}
%    \item Our \acrosskv can be shared for more than two layers. However, \citep{liu2024minicache} can be only applied to two adjacent layers.
%    \item \citep{liu2024minicache} has to retain certain critical tokens in every layer (referred to as Token Retention Strategy) but our \acrosskv does not have this requirement.
%\end{itemize}
%As such, 

%\zhewei{Our result is difference than MiniCache!!! Need @aurick to take a look!!! --- Aurick cannot find any issue... but after I did a loop to get the best match heads, I still cannot reproduce the result. I have some hypotenthesis; let me verify later.}

\subsection{Knowledge Recovery} 
\label{sec:knowledge-recovery}

While \singlekv preserves all the original parameters, it re-wires the architecture so that the \kv projections may receive different inputs. We found that this re-wiring (and \acrosskv) requires fine-tuning to recover the original capabilities from the modified model. Since we only change the KV projections for layer $> l$, this can be achieved by fine-tuning just the \( \rmW_{QKV} \) weight matrices from the $(l+1)$-th layer onwards. However, instead of directly fine-tuning these parameters using standard LM loss, we find that distilling using the output logits of the original model allows for better knowledge recovery (see \sref{sec:discussion} for more details).

Additionally, we found that limiting the training to just \( \rmW_{QKV} \) achieves better accuracy, which aligns with prior hypotheses that LLM knowledge is primarily stored in their MLP layers~\cite{10.5555/3600270.3601532,geva-etal-2021-transformer,elhage2021mathematical}. We further explore this in Sec.~\ref{sec:impact-of-distillation}. An added benefit is that these parameters are typically <10\% of the total for popular GQA models (e.g., Llama, Mistral, Qwen), allowing for very efficient distillation.

\paragraph{Efficient Distillation.} Since only a few \( \rmW_{QKV} \) parameters need training, we can keep just a single copy of the original model weights in memory that are frozen during training, and add an extra trainable copy of the \( \rmW_{QKV} \) parameters for layers $> l$ initialized using the original model (See \fref{fig:main-figure}). 

During training, we create two modes for the later layers $> l$, one with original frozen parameters using original architecture, and another with the \OURS re-wiring using new QKV projections i.e.,
\begin{equation}
\begin{aligned}
    \y_{teacher} &= \rmM(\x, \OURS=False), \\
    \y_{student} &= \rmM(\x, \OURS=True),
\end{aligned}
\end{equation}
where $\y_{\cdot}$ is the final logits, $\rmM$ is the model, and $\x$ is the input. Afterwards, we apply the standard distillation loss~\cite{journals/corr/HintonVD15} on the outputs. 
After the distillation, the original KV projection layers $> l$ are discarded during inference.
%{\color{blue} [YH:Does the inference model needs to keep two versions of weights?  If not, please clarify.] --> {\color{red} SR: Re-wrote most of this section, also added the clarification.}}
%\zhewei{In our main figure, we should have this component as well.}

% \zhewei{We should also have an ablation section to discuss the goodness of our distillation} Done

This method allows us to distill Llama-3.1-8B-Instruct on 680M tokens of data in 3 hours using 8 H100 GPUs, and Llama-3.1-70B-Instruct in 5 hours using 32 H100 GPUs across 4 nodes. In contrast, many prune-and-distill~\cite{sreenivas2024llmpruningdistillationpractice} and layer-skipping~\cite{elhoushi-etal-2024-layerskip} methods require much larger datasets (e.g. 10--100B tokens) and incur greater accuracy gaps than \OURS.

\subsection{Optimized Implementation for Inference}

LLM serving systems can be complex and incorporate many simultaneous optimizations at multiple layers of the stack, such as PagedAttention~\citep{10.1145/3600006.3613165}, Speculative Decoding~\citep{10.5555/3618408.3619203}, SplitFuse~\citep{holmes2024deepspeedfastgenhighthroughputtextgeneration,298679}, and more. A benefit of \OURS is that it makes minimal changes to the model architecture, so it can be integrated into existing serving systems without implementing new kernels (e.g. for custom attention operations or sparse computation) or novel inference procedures.

\paragraph{Implementation in vLLM and SGLang.} To show that the theoretical compute reductions of \OURS translates to real-world savings, we integrated it with vLLM~\cite{10.1145/3600006.3613165} and SGLang~\cite{zheng2024sglang}. Our implementation is compatible with chunked prefill~\cite{holmes2024deepspeedfastgenhighthroughputtextgeneration,298679}, which mixes chunks of prefill tokens and decode tokens in each minibatch. During each forward pass, after completing layer \( l \), the KV-cache for the remaining layers (\( >l \)) are immediately computed, and only the decode tokens are propagated through the rest of the model layers.

%\textbf{GEMM and Memory Optimizations.} Upon this basic implementation, we implemented two additional optimizations. First, instead of computing the \kv \(\rmKV_j \) for each layer \( j>l \) one at a time, we fused all $\rmW_{KV}^j$ into one large weight matrix $\rmW_{KV}^{j>l}$ so that their \kv can be computed with a single efficient GEMM operation. Second, \emph{\acrosskv reduction}: we modified vLLM to only allocate one layer's KV-cache for each group of merged layers, which realizes the memory gains of \acrosskv. 

%{\color{blue} [YH: What about the KV ache of the generated tokens?  They are actually calculated (instead of projected), right? Please clarify this part. } --> {\color{red} Samyam: Clarified. Only the decode tokens go though the rest of the layers but the KV cache for both have been computed at layer l.}

%\aurick{We can either add a diagram illustrating these, or a small table that ablates \singlekv fusion and \acrosskv reduction}
\section{Main Results}
\label{sec:main_results}

We evaluated \OURS in terms of model accuracy (\sref{subsec:model-quality-vs-compression}) compared to the original model and several baselines, and end-to-end inference performance (\sref{sec:inference-speedup}) in a real serving system.

%\subsection{Setup}

\paragraph{Distillation datasets.} Our dataset is a mixture of Ultrachat~\citep{ding2023enhancing}, SlimOrca~\citep{SlimOrca}, and OpenHermes-2.5~\citep{OpenHermes}, totaling roughly 680M Llama-3.1 tokens. 
%We evaluated model quality using a modified LM-Eval-Harness~\citep{eval-harness}\footnote{\scriptsize\url{https://github.com/neuralmagic/lm-evaluation-harness/tree/llama_3.1_instruct}} due to its support for the custom prompt format of \llama, particularly for MMLU and MMLU-CoT~\citep{hendryckstest2021}, GSM8K~\citep{cobbe2021training}, and Arc-Challenge~\citep{Clark2018ThinkYH}.
For more details, please see~\appref{sec:training_details}. %{\color{blue}[YH: what does it mean by high performance for Eval?]}\aurick{rewritten for clarity}
%We add two extra dataset, Winogrande and Helloswag, for more comparisons between base models and \OURS. 

\paragraph{\OURS Notation.} For prefill computation, we report the approximate reduction as $(L-l) / L$  due to \singlekv, and for \kv, we report the exact memory reduction due to \acrosskv. For example, \singlekv (\( l = L/2 \)) and 4-way \acrosskv is reported as 50\% prefill compute reduction and 37.5\% \kv memory reduction.
%We further study how these theoretical compute and memory reductions translate into end-to-end inference improvements in \sref{sec:inference-speedup}.
%{\color{red} [YH: check revised reduction formula, the original was $l/L$]}
%use the approximation reduction for the prefill phase for simplicity. Particularly, if $l=L/2$ in~\eqref{eq:singlekv}, then we approximate the reduction as 50\% as additional computation from \kv projection matrices, i.e., $\rmW_{kv}$s are small due to GQA \cite{ainslie-etal-2023-gqa}. For \kv reduction, we directly measure the real reduction ratio. 
%{\color{blue}[YH: Please massage the above 2 sentences. It reads a bit odd.]} \aurick{rewritten}

%\textbf{Why not comparing with other methods?} The most important aspect of \OURS is that it can directly reduce the prefill phase computation for enterprise usage. 
%However, existing methods are normally not in this category. 
%For example, compression of \kv can just reduce \kv size without reducing computation during the prefill phase.
%For other early existing methods, they may (or may not) skip the rest of layers for certain token generation. 
%But they have to recompute the later layers \kv when early existing fails. 
%As such, they defer the calculation instead of reducing the cost.
%Therefore, we do not compare \OURS with other methods based on the best of our knowledge.

\subsection{Model Quality Impact of \OURS}
\label{subsec:model-quality-vs-compression}

\begin{table*}[t]
\caption{
All \OURS model quality evaluations. For FFN-SkipLLM, we set the candidate layers to be skipped to be from 35--8\% depth in each model, which reflects the settings in their paper. The prefill reduction \% represents just the fraction of MLP layer skipped, and varies between models and tasks since it is adaptively determined during inference.
}\centering
\label{tab:main_result}
\begin{adjustbox}{width=\linewidth}
\centering
\begin{tabular}{llcccccccccccccc }
\toprule[2pt]
\multirow{2}{*}{Model} &  & \singlekv & \acrosskv  & Arc-Challenge  & Winogrande & Hellaswag & TruthfulQA & MMLU     & MMLU-CoT & GSM8K-CoT & \multirow{2}{*}{Avg.}\\
      &  & (Prefill Reduction) & (Cache Reduction)            & 0-shot         & 5-shot    & 10-shot & 0-shot & 5-shot  & 0-shot   & 8-shot \\
\midrule[2pt]
\multirow{9}{*}{Llama-3.1-8B-Instruct} & Baseline & -- & -- &82.00 &77.90 &80.40 &54.56 &67.90 &70.63 &82.56 &73.71 \\
\cmidrule{2-12}
& \OURS & \cmark (25\%) & \xmark &82.08 &77.98 &80.63 &54.59 &67.95 &70.45 &81.43 &73.59 \\
& \OURS & \cmark (50\%) & \xmark &80.38 &78.22 &79.30 &54.54 &67.30 &69.73 &79.45 &72.70 \\
& \OURS & \cmark (62.5\%) & \xmark &71.76 &75.77 &78.21 &52.73 &61.55 &53.68 &68.92 &66.09\\
% \OURS & \cmark (75\%) & \xmark &47.01 &69.38 &74.51 &44.09 &42.93 &24.23 &42.68 &49.26\\
\cmidrule{2-12}
% \OURS & \cmark (50\%) & \xmark \\
& \OURS & \cmark (50\%) & 2-way (25\%)~~~~~~~ &80.29 &77.82 &79.03 &54.66 &66.96 &68.39 &75.59 &71.82 \\
% \OURS & \cmark (50\%) & \cmark (31.25\%) \\
& \OURS & \cmark (50\%) & 4-way (37.5\%)~~~~ &79.35 &77.51 &78.44 &54.96 &65.71 &67.75 &76.72 &71.49 \\
& \OURS & \cmark (50\%) & 8-way (43.75\%)~~ &79.18 &77.19 &77.38 &54.79 &65.73 &66.88 &72.33 &70.50\\
& \OURS & \cmark (50\%) & 16-way (46.875\%) &78.24 &76.80 &76.87 &56.86 &64.65 &65.86 &72.25 &70.22\\
\cmidrule{2-12}
& FFN-SkipLLM & (12-19\%) & -- & 81.4 & 74.11 & 73.94 & 54.55 & 67.65 & 64.12 & 78.62 & 70.62 \\
\midrule[2pt]
\multirow{7}{*}{Llama-3.1-70B-Instruct} & Baseline & -- & -- &93.34 &85.16 &86.42 &59.95 &83.97 &86.21 &95.15 &84.31 \\
\cmidrule{2-12}
& \OURS & \cmark (25\%) & \xmark &93.00 &84.69 &85.98 &59.43 &82.82 &85.81 &95.07 &83.83 \\
& \OURS & \cmark (50\%) & \xmark &93.09 &83.82 &84.45 &58.40 &82.51 &85.00 &93.56 &82.98 \\
% \OURS & \cmark (62.5\%) & \xmark &71.76 &75.77 &78.21 &52.73 &61.55 &53.68 &68.92 &66.09\\
% \OURS & \cmark (75\%) & \xmark &47.01 &69.38 &74.51 &44.09 &42.93 &24.23 &42.68 &49.26\\
\cmidrule{2-12}
% \OURS & \cmark (50\%) & \xmark \\
& \OURS & \cmark (50\%) & 2-way (25\%) &92.92 &82.95 &84.10 &57.79 &82.66 &84.55 &93.48 &82.63 \\
% \OURS & \cmark (50\%) & \cmark (31.25\%) \\
& \OURS & \cmark (50\%) & 4-way (37.5\%) &92.92 &83.74 &84.72 &58.28 &82.60 &84.79 &93.71 &82.96\\
% \OURS & \cmark (50\%) & \cmark (43.75\%) &79.18 &77.19 &77.38 &54.79 &65.73 &66.88 &72.33 &70.50\\
\cmidrule{2-12}
& Nemotron-51B & (28\%) & (50\%) & 91.47 & 84.45 & 85.68 & 59.02 & 81.74 & 83.86 & 93.25 & 82.78 \\
\midrule[2pt]
\multirow{2}{*}{Llama-3.1-405B-Instruct (FP8)} & Baseline & -- & -- &94.7&87.0&88.3&64.7&87.5&88.1&96.1&86.6 \\
\cmidrule{2-12}
& \OURS & \cmark (50\%) & \xmark &94.0&86.3&88.1&64.2&85.7&87.5&95.2&85.9 \\
\midrule[2pt]
\multirow{8}{*}{Llama-3.2-3B-Instruct} & Baseline & -- & -- & 75.17 & 68.59 & 73.32 & 51.45 & 62.01 & 62.48 & 72.32 & 66.47 \\
\cmidrule{2-12}
& \OURS & \cmark (25\%) & \xmark &75.59&69.77&72.34&52.80&61.89&62.39&71.11&66.55 \\
& \OURS & \cmark (40\%) & \xmark &75.34&68.98&71.37&51.10&61.80&61.62&68.68&65.55 \\
& \OURS & \cmark (50\%) & \xmark &71.25&68.75&70.77&51.29&59.63&59.94&67.02&64.09 \\
\cmidrule{2-12}
& \OURS & \cmark (40\%) & 2-way (25\%) &74.82&68.66&71.41&50.67&61.55&61.03&67.77&65.13 \\
& \OURS & \cmark (40\%) & 4-way (37.5\%) &75.59&69.21&70.79&50.89&61.35&60.82&67.70&65.19 \\
\cmidrule{2-12}
& FFN-SkipLLM & (8-16\%) & -- & 74.57 & 66.38 & 67.55 & 49.57 & 60.95 & 61.24 & 69.75 & 64.28 \\
\midrule[2pt]
\multirow{7}{*}{Mistral-Small-Instruct-2409} & Baseline & -- & -- &84.12&84.68&87.27&56.85&73.33&74.86&86.50&78.23 \\
\cmidrule{2-12}
& \OURS & \cmark (25\%) & \xmark &84.04&84.84&87.03&55.97&72.88&74.69&85.21&77.80 \\
& \OURS & \cmark (50\%) & \xmark &83.53&83.97&86.30&55.63&72.91&74.04&84.30&77.24 \\
\cmidrule{2-12}
& \OURS & \cmark (50\%) & 2-way (25\%) &83.36&84.05&86.22&56.20&72.30&73.70&84.68&77.21 \\
& \OURS & \cmark (50\%) & 4-way (37.5\%) &82.93&83.82&86.17&56.00&72.29&73.00&82.48&76.66 \\
\cmidrule{2-12}
& FFN-SkipLLM & (34-37\%) & -- & 65.61 & 72.61 & 59.80 & 53.52 & 64.20 & 2.16 & 2.12 & 45.71 \\
\midrule[2pt]
\multirow{7}{*}{Deepseek-V2-Lite-Chat} & Baseline & -- & -- &65.53&74.66&81.56&50.98&56.86&50.61&68.69&64.12 \\
\cmidrule{2-12}
& \OURS & \cmark (25\%) & \xmark &65.44&75.05&81.52&50.53&56.91&50.92&68.99&64.19 \\
& \OURS & \cmark (45\%) & \xmark &65.61&73.95&80.82&50.20&56.33&51.56&66.11&63.51 \\
\cmidrule{2-12}
& \OURS & \cmark (45\%) & 2-way (25\%) &65.52&74.26&80.23&49.85&55.59&50.51&65.57&63.07 \\
& \OURS & \cmark (45\%) & 4-way (37.5\%) &61.34&75.21&79.80&48.39&54.82&30.80&64.89&59.32 \\
\cmidrule{2-12}
& FFN-SkipLLM & (30-32\%) & -- & 10.49 & 58.41 & 49.34 & 50.69 & 4.56 & 0.01 & 0.30 & 24.83 \\
\midrule[2pt]
\multirow{8}{*}{Qwen2.5-14B-Instruct} & Baseline & -- & -- & 62.29 & 79.32 & 85.04 & 69.07 & 76.58 & 79.04 & 90.37 & 77.38 \\
\cmidrule{2-12}
& \OURS & \cmark (25\%) & \xmark & 62.03 & 79.00 & 84.63 & 68.39 & 76.09 & 78.64 & 84.83 & 76.23 \\
& \OURS & \cmark (50\%) & \xmark & 56.91 & 77.26 & 82.71 & 60.76 & 64.40 & 68.20 & 79.30 & 69.93 \\
\cmidrule{2-12}
& \OURS & \cmark (25\%) & 2-way (25\%) & 61.43 & 79.71 & 85.22 & 69.33 & 76.25 & 78.88 & 84.23 & 76.43 \\
& \OURS & \cmark (25\%) & 4-way (37.5\%) & 59.13 & 80.89 & 84.92 & 68.75 & 75.70 & 78.84 & 82.78 & 75.85 \\
\cmidrule{2-12}
& FFN-SkipLLM & (7--21\%) & -- & 53.24 & 73.09 & 65.10 & 59.78 & 73.55 & 62.22 & 50.79 & 62.53 \\
& DarwinLM-8.4B & (40\%) & -- & 49.32 & 70.96 & 74.95 & 41.99 & 12.46 & 0.00 & 1.90 & 35.94 \\
% \midrule
% \multirow{6}{*}{Qwen2.5-32B-Instruct} & Baseline & -- & -- & - & 80.11 & 85.92 & 65.59 & 80.60 & 82.16 & 91.50 & - \\
% \cmidrule{2-12}
% & \OURS & \cmark (25\%) & \xmark & - & 79.48 & 85.48 & 65.77 & 80.63 & 82.36 & 90.90 & - \\
% & \OURS & \cmark (50\%) & \xmark & - & 75.37 & 83.11 & 60.83 & 61.04 & 64.27 & 81.95 & - \\
% \cmidrule{2-12}
% & \OURS & \cmark (25\%) & 2-way (25\%) & - & 79.55 & 85.48 & 65.33 & 80.50 & 82.06 & 90.59 & - \\
% & \OURS & \cmark (25\%) & 4-way (37.5\%) & - & 79.95 & 85.57 & 66.93 & 79.80 & 81.86 & 90.37 & - \\
% \cmidrule{2-12}
% & FFN-SkipLLM & (45-48\%) & -- & - & 52.64 & - & 49.01 & - & - & - & -\\
\bottomrule[2pt]
\end{tabular}
\end{adjustbox}
\end{table*}

\tref{tab:main_result} shows the quality results of all models we evaluated, including Llama-3.1-Instruct, Qwen2.5-14B-Instruct, Mistral-Small, and Deepseek-V2.
Of these models, we note that the Llama models span two orders of magnitude in size (3B to 405B), Llama-3.1-405B-Instruct uses FP8 (W8A16) quantization, and Deepseek-V2-Lite-Chat is a mixture-of-experts model that implements a novel latent attention mechanism~\citep{deepseekai2024deepseekv2strongeconomicalefficient}.

We also compare with three baselines: (1) \emph{FFN-SkipLLM}~\citep{jaiswal2024ffnskipllmhiddengemautoregressive}, a training-free method for skipping FFN layers (no attention layers are skipped) based on hidden state similarity, (2) \emph{Llama-3.1-Nemotron-51B-Instruct}~\cite{sreenivas2024llmpruningdistillationpractice}, which is pruned and distilled from Llama-3.1-70B-Instruct using neural architecture search on 40B tokens, and (3) \emph{DarwinLM-8.4B}~\cite{tang2025darwinlmevolutionarystructuredpruning}, which is pruned and distilled from Qwen2.5-14B-Instruct using 10B tokens.

\paragraph{\singlekv.} For Llama, Mistral, and Deepseek, we find the accuracy degradation for 25\% \OURS is less than 0.5\% from the original models (averaged across tasks). Additionally, the accuracy gap is within 1--2\% even at 40--50\% \OURS.
Beyond 50\% \singlekv, model quality drops quickly. For example, Llama-3.1-8B-Instruct incurs a 7\% accuracy gap at 62.5\% \singlekv.
%This can be explained by the drop in activation similarity from 0.61 to 0.51 between layer 16 to layer 12 (\fref{fig:motivation}).
We find that Qwen suffers larger degradations, at 1.1\% for 25\% \OURS and 7.4\% for 50\% \OURS, which may be due to Qwen models having lower simularity between layer at 50--75\% depth (\fref{fig:motivation}).
Even still, \OURS performs much better than FFN-SkipLLM and DarwinLM-8.4B, which suffer massive 15\% and 42\% drops from the baseline model, respectively.

\paragraph{\acrosskv.} The accuracy impact of \acrosskv is also minimal. Starting from 25--50\% \singlekv, adding 2-way \acrosskv (20--25\% \kv reduction) further degrades average task accuracy by at most 1\% across all models. Pushing to 4-way \acrosskv, Deepseek-V2-Lite-Chat experiences a steep accuracy drop from 63.07\% to 59.32\%, while other models experience smaller drops. Notably, we found that Llama-3.1-8B-Instruct still achieves 70.22\% average accuracy at 16-way \acrosskv, meaning all the last half of layers share a single layer of \kv.
Furthermore, the design of \acrosskv is complementary to many existing \kv compression methods. In \sref{subsec:kv-quantization}, we show that \acrosskv can be combined with quantization to achieve 62.5\% reduction in \kv memory.

\paragraph{\OURS vs Baselines.} \OURS outperforms FFN-SkipLLM across all scenarios we tested. FFN-SkipLLM skips only MLPs for prefill and decode tokens, while \OURS skips both MLP and attention layers for prefill tokens. Still, FFN-SkipLLM sees large degradations for Mistral, Deepseek, and Qwen, even at 7--37\% of MLPs skipped. For Llama models, skipping under 20\% of of the MLP layers using FFN-SkipLLM still underperforms \OURS skipping 50\% of MLP and attention layers.

Compared with Nemotron-51B and DarwinLM-8.4B, 50\% \OURS reduces \emph{more} prefill while achieving \emph{higher} accuracies. Also, Nemotron-51B is distilled on 40B tokens, and DarwinLM-8.4B on 10B tokens, while \OURS is distilled on <1B tokens and on <10\% of the model parameters. When prefill compute is substantial (e.g., many enterprise applications), \OURS is the clear choice for reducing cost without sacrificing accuracy.

\subsection{Inference Performance }
\label{sec:inference-speedup}

We focus on two common production scenarios:
\begin{enumerate}
\item \emph{Batch-Inference:} When processing requests in bulk or serving a model under high usage demand, it is important to achieve high \emph{combined throughput} in terms of input and output tokens to cost-effectively serve the model.

\item \emph{Interactive-Inference:} In interactive scenarios (e.g., chatbots, copilots), metrics that define the end-user experience are the time-to-first-token (TTFT) and time-per-output-token (TPOT).
%TTFT is the time between the user sending a message and receiving the first token in the response. TPOT is the time between each output token after the first token has been received.
Low TTFT and TPOT are desirable to deliver smooth usage experiences.
\end{enumerate}

We evaluate the end-to-end inference performance using \llamasmall running on 1 NVIDIA H100 GPU with 80GB of memory, \llamalarge running on 4 NVIDIA H100 GPUs with 4-way tensor parallelism. We show results using vLLM and refer to our SGLang results in \appref{sec:sglang-results}, and provide the full hardware and vLLM configurations in \appref{sec:speedup-details}.

\begin{figure}
\centering
\includegraphics[width=0.5\textwidth,trim={30 600 128 30},clip]{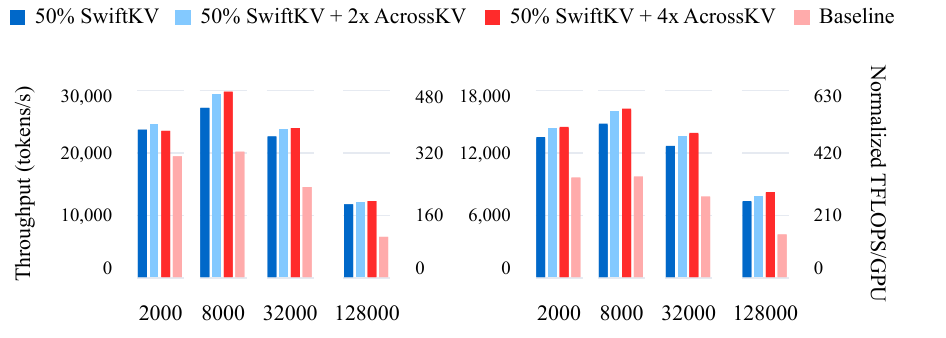}
\caption{
Combined input and output throughput for Llama-3.1-8B-Instruct (left) and Llama-3.1-70B-Instruct (right) across input lengths (bottom).
Roughly 15M tokens worth of requests are sent for each experiment, and each request generates 256 output tokens.
}
\label{fig:throughput}
\end{figure}

\paragraph{Batch Inference Performance.}
\fref{fig:throughput} shows the results of Llama-3.1-8B-Instruct and Llama-3.1-70B-Instruct across several workloads with a range of input lengths. \OURS achieves higher combined throughput than the baseline across all the workloads we evaluated. 
For \llamasmall, with 2K input tokens per prompt, \OURS achieves \(1.2-1.3\times\) higher combined throughput than the baseline, and our benefits increase further to \(1.8-1.9\times\) higher combined throughput with 128K inputs. Note that for an input length of 8K tokens, \OURS achieves a staggering 30K tokens/sec/GPU (480 TFLOPS/GPU). For Llama-3.1-70B-Instruct with 2K input tokens per prompt, \OURS achieves \(1.4-1.5\times\) higher combined throughput than the baseline, which improves to \(1.8-2.0\times\) better combined throughput for 128K inputs.
%\footnote{While the total compute savings is roughly \(2\times\), the end-to-end speedup is lower due to two main reasons: i) the performance improvement is limited to the decoding computation which needs the output activation of all the layers. Fig.~\ref{fig:motivation} (right) shows the  max possible speedup for \llamasmall during model forward pass despite the decoding overhead, and ii) due to additional vLLM overheads outside of the model forward pass, such as sampling, optimizing which is beyond the scope of the paper.} As expected, \OURS provides greater improvements when the inputs are long.

\begin{table}[t]
\caption{
Throughput of \llamasmall compared between Baseline, Merge-all-Layers, and SwiftKV variants. Run on a H100 GPU with varying memory limits.
}
\label{tab:compute_vs_memory}
\centering
\begin{adjustbox}{width=\linewidth}
\footnotesize
\begin{tabular}{lccccc}
\toprule
\multicolumn{6}{c}{Throughput (tokens/s)} \vspace{1mm} \\
\multirow{2}{*}{Memory} & \multirow{2}{*}{Baseline} & Merge-all- & \multirow{2}{*}{50\% \singlekv} & 50\% \singlekv & 50\% \singlekv  \\
 & & Layers & & + \(4\times\) AcrossKV & + \(4\times\) AcrossKV (FP8) \\
\midrule
80GB & 22.9K & 25.1K & 31.0K & 31.2K & 32.0K \\
40GB & 20.6K & 25.2K & 27.3K & 28.4K & 28.9K \\
20GB & 10.8K & 25.2K & 12.2K & 18.0K & 23.2K \\
16GB & OOM & 24.8K & OOM & 4.22K & 7.28K \\
\bottomrule
\end{tabular}
\end{adjustbox}
\end{table}

\begin{figure}
\centering
\includegraphics[width=0.5\textwidth,trim={20 550 120 20},clip]{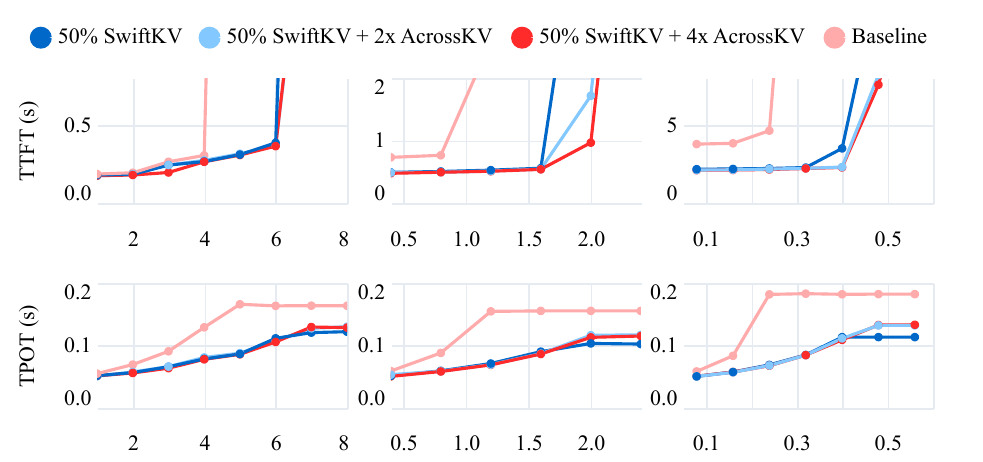}
\caption{
Time to first token (TTFT, top) and time per output token (TPOT, bottom) for input lengths 2000 (left), 8000 (middle), and 32000 (right) for Llama-3.1-70B. For each experiment, a range of different request arrival rates is simulated. Each request generates 256 output tokens.
}
\label{fig:latency-70b}
\end{figure}

We also observe \acrosskv can further improve the combined throughput due to its ability to reduce the memory usage for the KV-cache and supporting larger batch sizes. For sequence length of 8K, \llamalarge with \OURS achieves a combined throughput of over 16K toks/sec over 4xH100 GPUs which corresponds to 560 TFLOPS/GPU of BF16 performance when normalized to baseline. This is an unprecedented throughput for BF16 inference workloads. 

\paragraph{Interactive-Inference Performance.} 
\fref{fig:latency-70b} shows the TTFT and TPOT of \llamalarge across a range of request arrival rates and input lengths, and we refer to \fref{fig:latency-8b} in the Appendix for Llama-3.1-8B-Instruct. When the arrival rate is too high, the TTFT explodes due to the request queue accumulating faster than they can be processed by the system. However, \OURS can sustain \(1.5-2.0\times\) higher arrival rates before experiencing such TTFT explosion. When the arrival rate is low, \OURS can reduce the TTFT by up to 50\% for workloads with longer input lengths. In terms of TPOT, \OURS achieves significant reductions for all but the lowest arrival rates, up to 60\% for certain settings.

At first, it may be counter-intuitive that \OURS can reduce TPOT by only optimizing the prefill compute and not decode compute. However, in most open-source inference systems today, including vLLM and SGLang, prefill and decode are run on the same GPUs, whether they be interleaved~\cite{280922} or mixed~\cite{holmes2024deepspeedfastgenhighthroughputtextgeneration,298679}. This means prefill and decode may contend for GPU time, and reducing prefill compute also benefits decode latency.

\paragraph{Inference on Real-World Requests.}

In \appref{sec:sharegpt-results}, we evaluate \OURS on SGLang using real-world requests from ShareGPT~\cite{sharegpt}, which are collected in the wild from users of ChatGPT~\cite{chatgpt}. We show that the throughput improvements due to \OURS transfer well to real-world length distributions.

\section{Ablations and Discussions}
\label{sec:discussion}

\begin{table}
\caption{
Impact of Distillation and Full/Partial Model Finetuning on \llamasmall.
}
\label{tab:distillation_and_part_model}
\centering
\begin{adjustbox}{width=\linewidth}
\centering
\begin{tabular}{lcccccccccccccc }
\toprule
\multirow{2}{*}{Setting}  & Arc-Challenge  & Winogrande & Hellaswag & TruthfulQA & MMLU     & MMLU-CoT & GSM-8K & \multirow{2}{*}{Avg.}\\
              & 0-shot         & 5-shots    & 10-shots & 0-shot & 5-shots  & 0-shot   & 8-shots \\
\midrule
\multicolumn{9}{c}{(a) The effect of distillation}  \\
W/o Distill &79.44 &77.27 &78.71 &51.14 &65.55 &65.60 &72.71 &70.06 \\
W Distill &80.38 &78.22 &79.30 &54.54 &67.30 &69.73 &79.45 &72.70 \\
\midrule
\multicolumn{9}{c}{(b) Full model finetuning vs. part model finetuning} \\
\midrule
Full Model &76.79 &74.82 &76.42 &53.08 &62.94 &64.20 &69.37 &68.23 \\
Part Model &80.38 &78.22 &79.30 &54.54 &67.30 &69.73 &79.45 &72.70 \\
\bottomrule
\end{tabular}
\end{adjustbox}
\end{table}

\subsection{Compute vs Memory Reduction}
\label{sec:compute-vs-memory}

A key aspect of \OURS is combining prefill compute reduction and \kv compression (\acrosskv). While many prior works address \kv compression alone, they are only effective when GPU memory is limited, and are less impactful on datacenter GPUs (e.g., A100 and H100) with sufficient memory and inference is compute-bound.

To illustrate, we construct an ``ideal'' KV compression scheme, where every layer’s KV cache is merged into a single layer (Merge-all-Layers). We retain the computation for all KV operations (i.e., \(W_{kv}^TX\)) but eliminate the memory for all layers > 1, leading to a single layer of KV cache. Merge-all-Layers represents a “best case compression scenario” with (1) extreme compression ratio beyond any published technique, e.g. \(32\times\) and \(80\times\) for Llama-3.1 8B-Instruct and 70B-Instruct, respectively, and (2) zero overhead, while most techniques (e.g., quantization, low-rank decomposition) add extra computations or data conversions.

\tref{tab:compute_vs_memory} shows the throughput attained by Merge-all-Layers compared with the baseline model and its SwiftKV variants under various memory constraints. As shown, Merge-all-Layers outperforms only in very low memory scenarios (e.g. 16GB and 20GB) when there is barely enough memory for just the model weights, and is only marginally (10\%) better than the baseline model when using all 80GB memory. On the other hand, \OURS attains 35\% higher throughput than the baseline at 80GB even without \acrosskv. When combined with \( 4\times \) \acrosskv using FP8-quantized KV cache, \OURS can approach the throughput of Merge-all-Layers even at a more limited 20GB of memory.

\subsection{The Impact of Distillation}
\label{sec:impact-of-distillation}

To demonstrate the effectiveness of our distillation method, we train \llamasmall with 50\% \singlekv and no \acrosskv using the standard language model loss, and compare it with our distillation based approach discussed in ~\sref{sec:knowledge-recovery}. 
The results are shown in~\tref{tab:distillation_and_part_model} (a).
As we can see, the model trained with distillation has a 2.64 point higher average. 
Particularly, for generative tasks, i.e., MMLU-Cot and GSM-8K, the performance improvement is 4.13 and 6.74, respectively.

\paragraph{Full model training vs. partial model training.}

Our distillation method only fine-tuned the \( \rmW_{QKV} \) parameters
%, as discussed in~\sref{sec:knowledge-recovery}, 
hypothesizing that this preserves the original model's knowledge better than full model fine-tuning. This aligns with \cite{10.5555/3600270.3601532}, \cite{geva-etal-2021-transformer}, and \cite{elhage2021mathematical}, which suggest that MLP layers player a more prominent role in storing knowledge.

To validate this, we fine-tuned a model with 50\% \singlekv on \llamasmall where all parameters in the latter 50\% of layers are trained. 
The results are shown in~\tref{tab:distillation_and_part_model} (b). 
The model quality of full model distillation is about 4.5 points lower than our proposed partial model distillation.

\section{Conclusions}
\label{sec:conclusions}

We presented \OURS, a model transformation for reducing inference cost for prefill-dominant workloads, combined with a KV cache reduction strategy to reduce memory footprint, and a light-weight distillation procedure to preserve model accuracy. 
SwiftKV demonstrates strong results and leaves room for exploration in parameter-preserving transformations to further optimize inference.
%While we presented strong results on the effectiveness of \OURS, exploration of parameter-preserving model transformations for inference optimization is still in its early stages. We have identified both limitations as well as areas of improvement. Given the simplicity and effectiveness of \OURS, we hope that this will spark further exploration which we hope will continue to lower the latency and cost of inference.

% \section*{Limitations}

% %It is important for every work to acknowledge its limitations and suggest future directions, particularly for \llm-related works.
% In our work, we did not aim to optimize the  training data selection though we provide potential ways in~\sref{sec:dataset_impact}. 
% Additionally, we did not include a detailed benchmark analysis for our method.
% However, as shown in~\sref{sec:dataset_impact}, we ensured that our datasets were not cherry-picked to overfit the reported tasks.
% Furthermore, we did not finetune our model with advanced post-training approaches, like DPO and RLHF, which we leave for future work.
% Finally, we hypothesize that our method can work even better when combined with pretraining or continued-pretraining,
% but due to resources constraints, we did not explore this direction. 
% We hope to revisit these ideas in the future.

\bibliography{custom}

\begin{thebibliography}{56}
\providecommand{\natexlab}[1]{#1}

\bibitem[{Agrawal et~al.(2024)Agrawal, Kedia, Panwar, Mohan, Kwatra, Gulavani, Tumanov, and Ramjee}]{298679}
Amey Agrawal, Nitin Kedia, Ashish Panwar, Jayashree Mohan, Nipun Kwatra, Bhargav Gulavani, Alexey Tumanov, and Ramachandran Ramjee. 2024.
\newblock \href {https://www.usenix.org/conference/osdi24/presentation/agrawal} {Taming {Throughput-Latency} tradeoff in {LLM} inference with {Sarathi-Serve}}.
\newblock In \emph{18th USENIX Symposium on Operating Systems Design and Implementation (OSDI 24)}, pages 117--134, Santa Clara, CA. USENIX Association.

\bibitem[{Ainslie et~al.(2023{\natexlab{a}})Ainslie, Lee-Thorp, de~Jong, Zemlyanskiy, Lebron, and Sanghai}]{ainslie-etal-2023-gqa}
Joshua Ainslie, James Lee-Thorp, Michiel de~Jong, Yury Zemlyanskiy, Federico Lebron, and Sumit Sanghai. 2023{\natexlab{a}}.
\newblock \href {https://doi.org/10.18653/v1/2023.emnlp-main.298} {{GQA}: Training generalized multi-query transformer models from multi-head checkpoints}.
\newblock In \emph{Proceedings of the 2023 Conference on Empirical Methods in Natural Language Processing}, pages 4895--4901, Singapore. Association for Computational Linguistics.

\bibitem[{Ainslie et~al.(2023{\natexlab{b}})Ainslie, Lee-Thorp, de~Jong, Zemlyanskiy, Lebrón, and Sanghai}]{ainslie2023gqatraininggeneralizedmultiquery}
Joshua Ainslie, James Lee-Thorp, Michiel de~Jong, Yury Zemlyanskiy, Federico Lebrón, and Sumit Sanghai. 2023{\natexlab{b}}.
\newblock \href {https://arxiv.org/abs/2305.13245} {Gqa: Training generalized multi-query transformer models from multi-head checkpoints}.
\newblock \emph{Preprint}, arXiv:2305.13245.

\bibitem[{Ashkboos et~al.(2024)Ashkboos, Croci, do~Nascimento, Hoefler, and Hensman}]{ashkboos2024slicegptcompresslargelanguage}
Saleh Ashkboos, Maximilian~L. Croci, Marcelo~Gennari do~Nascimento, Torsten Hoefler, and James Hensman. 2024.
\newblock \href {https://arxiv.org/abs/2401.15024} {Slicegpt: Compress large language models by deleting rows and columns}.
\newblock \emph{Preprint}, arXiv:2401.15024.

\bibitem[{Chang et~al.(2024)Chang, Lin, Lin, Chen, Hu, Wang, Huang, Ceze, and Wu}]{chang2024palucompressingkvcachelowrank}
Chi-Chih Chang, Wei-Cheng Lin, Chien-Yu Lin, Chong-Yan Chen, Yu-Fang Hu, Pei-Shuo Wang, Ning-Chi Huang, Luis Ceze, and Kai-Chiang Wu. 2024.
\newblock \href {https://arxiv.org/abs/2407.21118} {Palu: Compressing kv-cache with low-rank projection}.
\newblock \emph{Preprint}, arXiv:2407.21118.

\bibitem[{Chen et~al.(2021)Chen, Tworek, Jun, Yuan, de~Oliveira~Pinto, Kaplan, Edwards, Burda, Joseph, Brockman, Ray, Puri, Krueger, Petrov, Khlaaf, Sastry, Mishkin, Chan, Gray, Ryder, Pavlov, Power, Kaiser, Bavarian, Winter, Tillet, Such, Cummings, Plappert, Chantzis, Barnes, Herbert-Voss, Guss, Nichol, Paino, Tezak, Tang, Babuschkin, Balaji, Jain, Saunders, Hesse, Carr, Leike, Achiam, Misra, Morikawa, Radford, Knight, Brundage, Murati, Mayer, Welinder, McGrew, Amodei, McCandlish, Sutskever, and Zaremba}]{chen2021evaluatinglargelanguagemodels}
Mark Chen, Jerry Tworek, Heewoo Jun, Qiming Yuan, Henrique~Ponde de~Oliveira~Pinto, Jared Kaplan, Harri Edwards, Yuri Burda, Nicholas Joseph, Greg Brockman, Alex Ray, Raul Puri, Gretchen Krueger, Michael Petrov, Heidy Khlaaf, Girish Sastry, Pamela Mishkin, Brooke Chan, Scott Gray, and 39 others. 2021.
\newblock \href {https://arxiv.org/abs/2107.03374} {Evaluating large language models trained on code}.
\newblock \emph{Preprint}, arXiv:2107.03374.

\bibitem[{Clark et~al.(2018)Clark, Cowhey, Etzioni, Khot, Sabharwal, Schoenick, and Tafjord}]{Clark2018ThinkYH}
Peter Clark, Isaac Cowhey, Oren Etzioni, Tushar Khot, Ashish Sabharwal, Carissa Schoenick, and Oyvind Tafjord. 2018.
\newblock Think you have solved question answering? try arc, the ai2 reasoning challenge.
\newblock \emph{ArXiv}, abs/1803.05457.

\bibitem[{Cobbe et~al.(2021)Cobbe, Kosaraju, Bavarian, Hilton, Nakano, Hesse, and Schulman}]{cobbe2021training}
Karl Cobbe, Vineet Kosaraju, Mohammad Bavarian, Jacob Hilton, Reiichiro Nakano, Christopher Hesse, and John Schulman. 2021.
\newblock \href {https://arxiv.org/abs/2110.14168} {Training verifiers to solve math word problems}.
\newblock \emph{Preprint}, arXiv:2110.14168.

\bibitem[{DeepSeek-AI et~al.(2024)DeepSeek-AI, Liu, Feng, Wang, Wang, Liu, Zhao, Dengr, Ruan, Dai, Guo, Yang, Chen, Ji, Li, Lin, Luo, Hao, Chen, Li, Zhang, Xu, Yang, Zhang, Ding, Xin, Gao, Li, Qu, Cai, Liang, Guo, Ni, Li, Chen, Yuan, Qiu, Song, Dong, Gao, Guan, Wang, Zhang, Xu, Xia, Zhao, Zhang, Li, Wang, Zhang, Zhang, Tang, Li, Tian, Huang, Wang, Zhang, Zhu, Chen, Du, Chen, Jin, Ge, Pan, Xu, Chen, Li, Lu, Zhou, Chen, Wu, Ye, Ma, Wang, Zhou, Yu, Zhou, Zheng, Wang, Pei, Yuan, Sun, Xiao, Zeng, An, Liu, Liang, Gao, Zhang, Li, Jin, Wang, Bi, Liu, Wang, Shen, Chen, Chen, Nie, Sun, Wang, Liu, Xie, Yu, Song, Zhou, Yang, Lu, Su, Wu, Li, Wei, Zhu, Xu, Huang, Li, Zhao, Sun, Li, Wang, Zheng, Zhang, Xiong, Zhao, He, Tang, Piao, Dong, Tan, Liu, Wang, Guo, Zhu, Wang, Zou, Zha, Ma, Yan, You, Liu, Ren, Ren, Sha, Fu, Huang, Zhang, Xie, Hao, Shao, Wen, Xu, Zhang, Li, Wang, Gu, Li, and Xie}]{deepseekai2024deepseekv2strongeconomicalefficient}
DeepSeek-AI, Aixin Liu, Bei Feng, Bin Wang, Bingxuan Wang, Bo~Liu, Chenggang Zhao, Chengqi Dengr, Chong Ruan, Damai Dai, Daya Guo, Dejian Yang, Deli Chen, Dongjie Ji, Erhang Li, Fangyun Lin, Fuli Luo, Guangbo Hao, Guanting Chen, and 138 others. 2024.
\newblock \href {https://arxiv.org/abs/2405.04434} {Deepseek-v2: A strong, economical, and efficient mixture-of-experts language model}.
\newblock \emph{Preprint}, arXiv:2405.04434.

\bibitem[{Ding et~al.(2023)Ding, Chen, Xu, Qin, Zheng, Hu, Liu, Sun, and Zhou}]{ding2023enhancing}
Ning Ding, Yulin Chen, Bokai Xu, Yujia Qin, Zhi Zheng, Shengding Hu, Zhiyuan Liu, Maosong Sun, and Bowen Zhou. 2023.
\newblock \href {https://arxiv.org/abs/2305.14233} {Enhancing chat language models by scaling high-quality instructional conversations}.
\newblock \emph{Preprint}, arXiv:2305.14233.

\bibitem[{Elhage et~al.(2021)Elhage, Nanda, Olsson, Henighan, Joseph, Mann, Askell, Bai, Chen, Conerly, DasSarma, Drain, Ganguli, Hatfield-Dodds, Hernandez, Jones, Kernion, Lovitt, Ndousse, Amodei, Brown, Clark, Kaplan, McCandlish, and Olah}]{elhage2021mathematical}
Nelson Elhage, Neel Nanda, Catherine Olsson, Tom Henighan, Nicholas Joseph, Ben Mann, Amanda Askell, Yuntao Bai, Anna Chen, Tom Conerly, Nova DasSarma, Dawn Drain, Deep Ganguli, Zac Hatfield-Dodds, Danny Hernandez, Andy Jones, Jackson Kernion, Liane Lovitt, Kamal Ndousse, and 6 others. 2021.
\newblock A mathematical framework for transformer circuits.
\newblock \emph{Transformer Circuits Thread}.
\newblock Https://transformer-circuits.pub/2021/framework/index.html.

\bibitem[{Elhoushi et~al.(2024)Elhoushi, Shrivastava, Liskovich, Hosmer, Wasti, Lai, Mahmoud, Acun, Agarwal, Roman, Aly, Chen, and Wu}]{elhoushi-etal-2024-layerskip}
Mostafa Elhoushi, Akshat Shrivastava, Diana Liskovich, Basil Hosmer, Bram Wasti, Liangzhen Lai, Anas Mahmoud, Bilge Acun, Saurabh Agarwal, Ahmed Roman, Ahmed Aly, Beidi Chen, and Carole-Jean Wu. 2024.
\newblock \href {https://doi.org/10.18653/v1/2024.acl-long.681} {{L}ayer{S}kip: Enabling early exit inference and self-speculative decoding}.
\newblock In \emph{Proceedings of the 62nd Annual Meeting of the Association for Computational Linguistics (Volume 1: Long Papers)}, pages 12622--12642, Bangkok, Thailand. Association for Computational Linguistics.

\bibitem[{Geva et~al.(2021)Geva, Schuster, Berant, and Levy}]{geva-etal-2021-transformer}
Mor Geva, Roei Schuster, Jonathan Berant, and Omer Levy. 2021.
\newblock \href {https://doi.org/10.18653/v1/2021.emnlp-main.446} {Transformer feed-forward layers are key-value memories}.
\newblock In \emph{Proceedings of the 2021 Conference on Empirical Methods in Natural Language Processing}, pages 5484--5495, Online and Punta Cana, Dominican Republic. Association for Computational Linguistics.

\bibitem[{GretelAI(2024)}]{gretelai_gsm8k_synthetic}
GretelAI. 2024.
\newblock Synthetically generated reasoning dataset (gsm8k-inspired) with enhanced diversity using gretel navigator and meta-llama/meta-llama-3.1-405b.
\newblock https://huggingface.co/gretelai/synthetic-gsm8k-reflection-405b.

\bibitem[{Gromov et~al.(2024)Gromov, Tirumala, Shapourian, Glorioso, and Roberts}]{gromov2024unreasonableineffectivenessdeeperlayers}
Andrey Gromov, Kushal Tirumala, Hassan Shapourian, Paolo Glorioso, and Daniel~A. Roberts. 2024.
\newblock \href {https://arxiv.org/abs/2403.17887} {The unreasonable ineffectiveness of the deeper layers}.
\newblock \emph{Preprint}, arXiv:2403.17887.

\bibitem[{Hendrycks et~al.(2021)Hendrycks, Burns, Basart, Zou, Mazeika, Song, and Steinhardt}]{hendryckstest2021}
Dan Hendrycks, Collin Burns, Steven Basart, Andy Zou, Mantas Mazeika, Dawn Song, and Jacob Steinhardt. 2021.
\newblock Measuring massive multitask language understanding.
\newblock \emph{Proceedings of the International Conference on Learning Representations (ICLR)}.

\bibitem[{Hinton et~al.(2015)Hinton, Vinyals, and Dean}]{journals/corr/HintonVD15}
Geoffrey~E. Hinton, Oriol Vinyals, and Jeffrey Dean. 2015.
\newblock \href {http://dblp.uni-trier.de/db/journals/corr/corr1503.html#HintonVD15} {Distilling the knowledge in a neural network.}
\newblock \emph{CoRR}, abs/1503.02531.

\bibitem[{Holmes et~al.(2024)Holmes, Tanaka, Wyatt, Awan, Rasley, Rajbhandari, Aminabadi, Qin, Bakhtiari, Kurilenko, and He}]{holmes2024deepspeedfastgenhighthroughputtextgeneration}
Connor Holmes, Masahiro Tanaka, Michael Wyatt, Ammar~Ahmad Awan, Jeff Rasley, Samyam Rajbhandari, Reza~Yazdani Aminabadi, Heyang Qin, Arash Bakhtiari, Lev Kurilenko, and Yuxiong He. 2024.
\newblock \href {https://arxiv.org/abs/2401.08671} {Deepspeed-fastgen: High-throughput text generation for llms via mii and deepspeed-inference}.
\newblock \emph{Preprint}, arXiv:2401.08671.

\bibitem[{Hooper et~al.(2024)Hooper, Kim, Mohammadzadeh, Mahoney, Shao, Keutzer, and Gholami}]{hooper2024kvquant10millioncontext}
Coleman Hooper, Sehoon Kim, Hiva Mohammadzadeh, Michael~W. Mahoney, Yakun~Sophia Shao, Kurt Keutzer, and Amir Gholami. 2024.
\newblock \href {https://arxiv.org/abs/2401.18079} {Kvquant: Towards 10 million context length llm inference with kv cache quantization}.
\newblock \emph{Preprint}, arXiv:2401.18079.

\bibitem[{Jaiswal et~al.(2024)Jaiswal, Hu, Yin, Ro, Liu, Chen, and Akella}]{jaiswal2024ffnskipllmhiddengemautoregressive}
Ajay Jaiswal, Bodun Hu, Lu~Yin, Yeonju Ro, Shiwei Liu, Tianlong Chen, and Aditya Akella. 2024.
\newblock \href {https://arxiv.org/abs/2404.03865} {Ffn-skipllm: A hidden gem for autoregressive decoding with adaptive feed forward skipping}.
\newblock \emph{Preprint}, arXiv:2404.03865.

\bibitem[{Jiang et~al.(2023)Jiang, Sablayrolles, Mensch, Bamford, Chaplot, de~las Casas, Bressand, Lengyel, Lample, Saulnier, Lavaud, Lachaux, Stock, Scao, Lavril, Wang, Lacroix, and Sayed}]{jiang2023mistral7b}
Albert~Q. Jiang, Alexandre Sablayrolles, Arthur Mensch, Chris Bamford, Devendra~Singh Chaplot, Diego de~las Casas, Florian Bressand, Gianna Lengyel, Guillaume Lample, Lucile Saulnier, Lélio~Renard Lavaud, Marie-Anne Lachaux, Pierre Stock, Teven~Le Scao, Thibaut Lavril, Thomas Wang, Timothée Lacroix, and William~El Sayed. 2023.
\newblock \href {https://arxiv.org/abs/2310.06825} {Mistral 7b}.
\newblock \emph{Preprint}, arXiv:2310.06825.

\bibitem[{Jiang et~al.(2024)Jiang, Li, Zhang, Wu, Luo, Ahn, Han, Abdi, Li, Lin, Yang, and Qiu}]{jiang2024minference10acceleratingprefilling}
Huiqiang Jiang, Yucheng Li, Chengruidong Zhang, Qianhui Wu, Xufang Luo, Surin Ahn, Zhenhua Han, Amir~H. Abdi, Dongsheng Li, Chin-Yew Lin, Yuqing Yang, and Lili Qiu. 2024.
\newblock \href {https://arxiv.org/abs/2407.02490} {Minference 1.0: Accelerating pre-filling for long-context llms via dynamic sparse attention}.
\newblock \emph{Preprint}, arXiv:2407.02490.

\bibitem[{Kwon et~al.(2023)Kwon, Li, Zhuang, Sheng, Zheng, Yu, Gonzalez, Zhang, and Stoica}]{10.1145/3600006.3613165}
Woosuk Kwon, Zhuohan Li, Siyuan Zhuang, Ying Sheng, Lianmin Zheng, Cody~Hao Yu, Joseph Gonzalez, Hao Zhang, and Ion Stoica. 2023.
\newblock \href {https://doi.org/10.1145/3600006.3613165} {Efficient memory management for large language model serving with pagedattention}.
\newblock In \emph{Proceedings of the 29th Symposium on Operating Systems Principles}, SOSP '23, page 611–626, New York, NY, USA. Association for Computing Machinery.

\bibitem[{Leviathan et~al.(2023)Leviathan, Kalman, and Matias}]{10.5555/3618408.3619203}
Yaniv Leviathan, Matan Kalman, and Yossi Matias. 2023.
\newblock Fast inference from transformers via speculative decoding.
\newblock In \emph{Proceedings of the 40th International Conference on Machine Learning}, ICML'23. JMLR.org.

\bibitem[{Lewis et~al.(2020)Lewis, Perez, Piktus, Petroni, Karpukhin, Goyal, K\"{u}ttler, Lewis, Yih, Rockt\"{a}schel, Riedel, and Kiela}]{10.5555/3495724.3496517}
Patrick Lewis, Ethan Perez, Aleksandra Piktus, Fabio Petroni, Vladimir Karpukhin, Naman Goyal, Heinrich K\"{u}ttler, Mike Lewis, Wen-tau Yih, Tim Rockt\"{a}schel, Sebastian Riedel, and Douwe Kiela. 2020.
\newblock Retrieval-augmented generation for knowledge-intensive nlp tasks.
\newblock In \emph{Proceedings of the 34th International Conference on Neural Information Processing Systems}, NIPS '20, Red Hook, NY, USA. Curran Associates Inc.

\bibitem[{Lian et~al.(2023)Lian, Wang, Goodson, Pentland, Cook, Vong, and "Teknium"}]{SlimOrca}
Wing Lian, Guan Wang, Bleys Goodson, Eugene Pentland, Austin Cook, Chanvichet Vong, and "Teknium". 2023.
\newblock \href {https://https://huggingface.co/Open-Orca/SlimOrca} {Slimorca: An open dataset of gpt-4 augmented flan reasoning traces, with verification}.

\bibitem[{Lin et~al.(2022)Lin, Hilton, and Evans}]{lin-etal-2022-truthfulqa}
Stephanie Lin, Jacob Hilton, and Owain Evans. 2022.
\newblock \href {https://doi.org/10.18653/v1/2022.acl-long.229} {{T}ruthful{QA}: Measuring how models mimic human falsehoods}.
\newblock In \emph{Proceedings of the 60th Annual Meeting of the Association for Computational Linguistics (Volume 1: Long Papers)}, pages 3214--3252, Dublin, Ireland. Association for Computational Linguistics.

\bibitem[{Lin et~al.(2024)Lin, Chen, Chen, Shi, Lomeli, James, Rodriguez, Kahn, Szilvasy, Lewis, Zettlemoyer, and tau Yih}]{lin2024radit}
Xi~Victoria Lin, Xilun Chen, Mingda Chen, Weijia Shi, Maria Lomeli, Richard James, Pedro Rodriguez, Jacob Kahn, Gergely Szilvasy, Mike Lewis, Luke Zettlemoyer, and Wen tau Yih. 2024.
\newblock \href {https://openreview.net/forum?id=22OTbutug9} {{RA}-{DIT}: Retrieval-augmented dual instruction tuning}.
\newblock In \emph{The Twelfth International Conference on Learning Representations}.

\bibitem[{Liu et~al.(2024{\natexlab{a}})Liu, Liu, Pan, He, Haffari, and Zhuang}]{liu2024minicache}
Akide Liu, Jing Liu, Zizheng Pan, Yefei He, Gholamreza Haffari, and Bohan Zhuang. 2024{\natexlab{a}}.
\newblock Minicache: Kv cache compression in depth dimension for large language models.
\newblock \emph{arXiv preprint arXiv:2405.14366}.

\bibitem[{Liu et~al.(2024{\natexlab{b}})Liu, Zeng, Li, Yan, Fu, Mei, and Chen}]{liu2024foldgptsimpleeffectivelarge}
Songwei Liu, Chao Zeng, Lianqiang Li, Chenqian Yan, Lean Fu, Xing Mei, and Fangmin Chen. 2024{\natexlab{b}}.
\newblock \href {https://arxiv.org/abs/2407.00928} {Foldgpt: Simple and effective large language model compression scheme}.
\newblock \emph{Preprint}, arXiv:2407.00928.

\bibitem[{Liu et~al.(2024{\natexlab{c}})Liu, Desai, Liao, Wang, Xie, Xu, Kyrillidis, and Shrivastava}]{10.5555/3666122.3668401}
Zichang Liu, Aditya Desai, Fangshuo Liao, Weitao Wang, Victor Xie, Zhaozhuo Xu, Anastasios Kyrillidis, and Anshumali Shrivastava. 2024{\natexlab{c}}.
\newblock Scissorhands: exploiting the persistence of importance hypothesis for llm kv cache compression at test time.
\newblock In \emph{Proceedings of the 37th International Conference on Neural Information Processing Systems}, NIPS '23, Red Hook, NY, USA. Curran Associates Inc.

\bibitem[{Ma et~al.(2023)Ma, Fang, and Wang}]{ma2023llmpruner}
Xinyin Ma, Gongfan Fang, and Xinchao Wang. 2023.
\newblock \href {https://openreview.net/forum?id=J8Ajf9WfXP} {{LLM}-pruner: On the structural pruning of large language models}.
\newblock In \emph{Thirty-seventh Conference on Neural Information Processing Systems}.

\bibitem[{Men et~al.(2024)Men, Xu, Zhang, Wang, Lin, Lu, Han, and Chen}]{men2024shortgptlayerslargelanguage}
Xin Men, Mingyu Xu, Qingyu Zhang, Bingning Wang, Hongyu Lin, Yaojie Lu, Xianpei Han, and Weipeng Chen. 2024.
\newblock \href {https://arxiv.org/abs/2403.03853} {Shortgpt: Layers in large language models are more redundant than you expect}.
\newblock \emph{Preprint}, arXiv:2403.03853.

\bibitem[{Meng et~al.(2024)Meng, Bau, Andonian, and Belinkov}]{10.5555/3600270.3601532}
Kevin Meng, David Bau, Alex Andonian, and Yonatan Belinkov. 2024.
\newblock Locating and editing factual associations in gpt.
\newblock In \emph{Proceedings of the 36th International Conference on Neural Information Processing Systems}, NIPS '22, Red Hook, NY, USA. Curran Associates Inc.

\bibitem[{OpenAI(2022)}]{chatgpt}
OpenAI. 2022.
\newblock \href {https://openai.com/blog/chatgpt} {[link]}.

\bibitem[{Pourreza and Rafiei(2024)}]{10.5555/3666122.3667699}
Mohammadreza Pourreza and Davood Rafiei. 2024.
\newblock Din-sql: decomposed in-context learning of text-to-sql with self-correction.
\newblock In \emph{Proceedings of the 37th International Conference on Neural Information Processing Systems}, NIPS '23, Red Hook, NY, USA. Curran Associates Inc.

\bibitem[{Pu et~al.(2023)Pu, Gao, and Wan}]{pu2023summarizationalmostdead}
Xiao Pu, Mingqi Gao, and Xiaojun Wan. 2023.
\newblock \href {https://arxiv.org/abs/2309.09558} {Summarization is (almost) dead}.
\newblock \emph{Preprint}, arXiv:2309.09558.

\bibitem[{Sakaguchi et~al.(2019)Sakaguchi, Bras, Bhagavatula, and Choi}]{sakaguchi2019winogrande}
Keisuke Sakaguchi, Ronan~Le Bras, Chandra Bhagavatula, and Yejin Choi. 2019.
\newblock Winogrande: An adversarial winograd schema challenge at scale.
\newblock \emph{arXiv preprint arXiv:1907.10641}.

\bibitem[{Schick et~al.(2023)Schick, Dwivedi-Yu, Dessì, Raileanu, Lomeli, Zettlemoyer, Cancedda, and Scialom}]{schick2023toolformerlanguagemodelsteach}
Timo Schick, Jane Dwivedi-Yu, Roberto Dessì, Roberta Raileanu, Maria Lomeli, Luke Zettlemoyer, Nicola Cancedda, and Thomas Scialom. 2023.
\newblock \href {https://arxiv.org/abs/2302.04761} {Toolformer: Language models can teach themselves to use tools}.
\newblock \emph{Preprint}, arXiv:2302.04761.

\bibitem[{{ShareGPT Team}(2023)}]{sharegpt}
{ShareGPT Team}. 2023.
\newblock \href {https://sharegpt.com/} {[link]}.

\bibitem[{Shazeer(2019)}]{shazeer2019fasttransformerdecodingwritehead}
Noam Shazeer. 2019.
\newblock \href {https://arxiv.org/abs/1911.02150} {Fast transformer decoding: One write-head is all you need}.
\newblock \emph{Preprint}, arXiv:1911.02150.

\bibitem[{Sreenivas et~al.(2024)Sreenivas, Muralidharan, Joshi, Chochowski, Patwary, Shoeybi, Catanzaro, Kautz, and Molchanov}]{sreenivas2024llmpruningdistillationpractice}
Sharath~Turuvekere Sreenivas, Saurav Muralidharan, Raviraj Joshi, Marcin Chochowski, Mostofa Patwary, Mohammad Shoeybi, Bryan Catanzaro, Jan Kautz, and Pavlo Molchanov. 2024.
\newblock \href {https://arxiv.org/abs/2408.11796} {Llm pruning and distillation in practice: The minitron approach}.
\newblock \emph{Preprint}, arXiv:2408.11796.

\bibitem[{Tang et~al.(2025)Tang, Sieberling, Kurtic, Shen, and Alistarh}]{tang2025darwinlmevolutionarystructuredpruning}
Shengkun Tang, Oliver Sieberling, Eldar Kurtic, Zhiqiang Shen, and Dan Alistarh. 2025.
\newblock \href {https://arxiv.org/abs/2502.07780} {Darwinlm: Evolutionary structured pruning of large language models}.
\newblock \emph{Preprint}, arXiv:2502.07780.

\bibitem[{Taori et~al.(2023)Taori, Gulrajani, Zhang, Dubois, Li, Guestrin, Liang, and Hashimoto}]{alpaca}
Rohan Taori, Ishaan Gulrajani, Tianyi Zhang, Yann Dubois, Xuechen Li, Carlos Guestrin, Percy Liang, and Tatsunori~B. Hashimoto. 2023.
\newblock Stanford alpaca: An instruction-following llama model.
\newblock \url{https://github.com/tatsu-lab/stanford_alpaca}.

\bibitem[{Teknium(2023)}]{OpenHermes}
Teknium. 2023.
\newblock \href {https://huggingface.co/datasets/teknium/OpenHermes-2.5} {Openhermes 2.5: An open dataset of synthetic data for generalist llm assistants}.

\bibitem[{Vaswani et~al.(2017)Vaswani, Shazeer, Parmar, Uszkoreit, Jones, Gomez, Kaiser, and Polosukhin}]{10.5555/3295222.3295349}
Ashish Vaswani, Noam Shazeer, Niki Parmar, Jakob Uszkoreit, Llion Jones, Aidan~N. Gomez, \L{}ukasz Kaiser, and Illia Polosukhin. 2017.
\newblock Attention is all you need.
\newblock In \emph{Proceedings of the 31st International Conference on Neural Information Processing Systems}, NIPS'17, page 6000–6010, Red Hook, NY, USA. Curran Associates Inc.

\bibitem[{Wang et~al.(2024)Wang, Wang, Athiwaratkun, Zhang, and Zou}]{wang2024mixtureofagentsenhanceslargelanguage}
Junlin Wang, Jue Wang, Ben Athiwaratkun, Ce~Zhang, and James Zou. 2024.
\newblock \href {https://arxiv.org/abs/2406.04692} {Mixture-of-agents enhances large language model capabilities}.
\newblock \emph{Preprint}, arXiv:2406.04692.

\bibitem[{Wei et~al.(2023)Wei, Wang, Liu, Ding, and Zhang}]{wei2023magicoder}
Yuxiang Wei, Zhe Wang, Jiawei Liu, Yifeng Ding, and Lingming Zhang. 2023.
\newblock Magicoder: Source code is all you need.
\newblock \emph{arXiv preprint arXiv:2312.02120}.

\bibitem[{Xia et~al.(2024)Xia, Gao, Zeng, and Chen}]{xia2024shearedllamaacceleratinglanguage}
Mengzhou Xia, Tianyu Gao, Zhiyuan Zeng, and Danqi Chen. 2024.
\newblock \href {https://arxiv.org/abs/2310.06694} {Sheared llama: Accelerating language model pre-training via structured pruning}.
\newblock \emph{Preprint}, arXiv:2310.06694.

\bibitem[{Yang et~al.(2024)Yang, Cao, and Zhao}]{yang2024lacolargelanguagemodel}
Yifei Yang, Zouying Cao, and Hai Zhao. 2024.
\newblock \href {https://arxiv.org/abs/2402.11187} {Laco: Large language model pruning via layer collapse}.
\newblock \emph{Preprint}, arXiv:2402.11187.

\bibitem[{Yao et~al.(2022)Yao, Aminabadi, Zhang, Wu, Li, and He}]{yao2022zeroquantefficientaffordableposttraining}
Zhewei Yao, Reza~Yazdani Aminabadi, Minjia Zhang, Xiaoxia Wu, Conglong Li, and Yuxiong He. 2022.
\newblock \href {https://arxiv.org/abs/2206.01861} {Zeroquant: Efficient and affordable post-training quantization for large-scale transformers}.
\newblock \emph{Preprint}, arXiv:2206.01861.

\bibitem[{Yu et~al.(2022)Yu, Jeong, Kim, Kim, and Chun}]{280922}
Gyeong-In Yu, Joo~Seong Jeong, Geon-Woo Kim, Soojeong Kim, and Byung-Gon Chun. 2022.
\newblock \href {https://www.usenix.org/conference/osdi22/presentation/yu} {Orca: A distributed serving system for {Transformer-Based} generative models}.
\newblock In \emph{16th USENIX Symposium on Operating Systems Design and Implementation (OSDI 22)}, pages 521--538, Carlsbad, CA. USENIX Association.

\bibitem[{Zellers et~al.(2019)Zellers, Holtzman, Bisk, Farhadi, and Choi}]{zellers2019hellaswag}
Rowan Zellers, Ari Holtzman, Yonatan Bisk, Ali Farhadi, and Yejin Choi. 2019.
\newblock Hellaswag: Can a machine really finish your sentence?
\newblock In \emph{Proceedings of the 57th Annual Meeting of the Association for Computational Linguistics}.

\bibitem[{Zhang et~al.(2024)Zhang, Ladhak, Durmus, Liang, McKeown, and Hashimoto}]{zhang-etal-2024-benchmarking}
Tianyi Zhang, Faisal Ladhak, Esin Durmus, Percy Liang, Kathleen McKeown, and Tatsunori~B. Hashimoto. 2024.
\newblock \href {https://doi.org/10.1162/tacl_a_00632} {Benchmarking large language models for news summarization}.
\newblock \emph{Transactions of the Association for Computational Linguistics}, 12:39--57.

\bibitem[{Zhao et~al.(2024)Zhao, Wu, and Wang}]{zhao2024alisaacceleratinglargelanguage}
Youpeng Zhao, Di~Wu, and Jun Wang. 2024.
\newblock \href {https://arxiv.org/abs/2403.17312} {Alisa: Accelerating large language model inference via sparsity-aware kv caching}.
\newblock \emph{Preprint}, arXiv:2403.17312.

\bibitem[{Zheng et~al.(2024)Zheng, Yin, Xie, Sun, Huang, Yu, Cao, Kozyrakis, Stoica, Gonzalez, Barrett, and Sheng}]{zheng2024sglang}
Lianmin Zheng, Liangsheng Yin, Zhiqiang Xie, Chuyue Sun, Jeff Huang, Cody~Hao Yu, Shiyi Cao, Christos Kozyrakis, Ion Stoica, Joseph~E. Gonzalez, Clark Barrett, and Ying Sheng. 2024.
\newblock \href {https://openreview.net/forum?id=VqkAKQibpq} {{SGL}ang: Efficient execution of structured language model programs}.
\newblock In \emph{The Thirty-eighth Annual Conference on Neural Information Processing Systems}.

\end{thebibliography}

\appendix
\onecolumn

\appendix

%%%%%%%%%%%%%%%%%%%
% Re-count the Figure/Algorithm/Tables after this point. 
%%%%%%%%%%%%%%%%%%%
\counterwithin{figure}{section}
\counterwithin{table}{section}

\section{Main Experiment Details}

\subsection{Training and Quality Evaluation Details}
\label{sec:training_details}

For datasets, we use a mixture of \texttt{HuggingFaceH4/ultrachat\_200k}, \texttt{teknium/OpenHermes-2.5}, and \texttt{Open-Orca/SlimOrca} which totals around 680M tokens. 
We set training epochs to be 2, learning rate to be 3e-4, weight decay to be 0.05, warm up ratio to be 5\%, maximum sequence length to be 8192 with attention separated sequence packing, the distillation temperature to be 2.0.

Our evaluation follows \url{https://huggingface.co/neuralmagic/Meta-Llama-3.1-8B-Instruct-FP8} using the github repository \url{https://github.com/neuralmagic/lm-evaluation-harness/tree/llama_3.1_instruct}. 
The main reason behind this is that the implementation implemented chat-templated evaluations for several of our evaluation tasks, which is especially important for the Llama-3.1/3.2 models. 
% One issue we found in the provided commands is the one used to run MMLU-5-shots. 
% Directly using the command does not give us desired accuracy.
% Therefore, we added both \texttt{-{}-apply\_chat\_template} and \texttt{-{}-fewshot\_as\_multiturn}, and the accuracy is even slightly higher than what they reported. 
For all tasks, we follow the same number of few shots and/or chain of thoughts as the provided commands. 
We present the number of shots and metrics used in the paper in~\tref{tab:metrics}.

\begin{table}
\caption{
The setting for different tasks
}
\label{tab:metrics}
\centering
\begin{adjustbox}{width=0.9\linewidth}
\centering
\begin{tabular}{cccccccccccccc}
\toprule
& Arc-Challenge  & Winogrande & HelloSwag & truthfulqa & MMLU     & MMLU-CoT & GSM-8K \\
\midrule
  & 0-shot         & 5-shots    & 10-shots & 0-shot & 5-shots  & 0-shot   & 8-shots \\
  \midrule
 & exact\_match,multi\_choice & acc & acc\_norm & truthfulqa\_mc2 (acc) & exact\_match,multi\_choice & exact\_match,strict-match & exact\_match,strict-match\\
\bottomrule
\end{tabular}
\end{adjustbox}
\end{table}

\subsection{Inference Speedup Evaluation Details}
\label{sec:speedup-details}

\paragraph{Hardware Details.} We ran all inference speedup experiments on a AWS p5.48xlarge instance, with 8 NVIDIA H100 GPUs, 192 vCPUs, and 2TB memory. \llamasmall experiments are run using 1 of the 8 GPUs, and \llamalarge experiments are run using 4 of the 8 GPUs.

\paragraph{vLLM Configuration.} We ran all experiments with \texttt{enforce\_eager} and chunked prefill enabled with \texttt{max\_num\_batched\_tokens} set to 2048. To run each benchmark, we instantiated vLLM's \texttt{AsyncLLMEngine} and submitted requests using its \texttt{generate} method according to each benchmark setting. For each request, the inputs are tokenized before being submitted, and the outputs are forced to a fixed length of 256.

\subsection{Llama-3.1-8B-Instruct Latency Results}
\label{sec:latency-8b}
\begin{figure}
\centering
\includegraphics[width=0.8\textwidth,trim={20 500 20 20},clip]{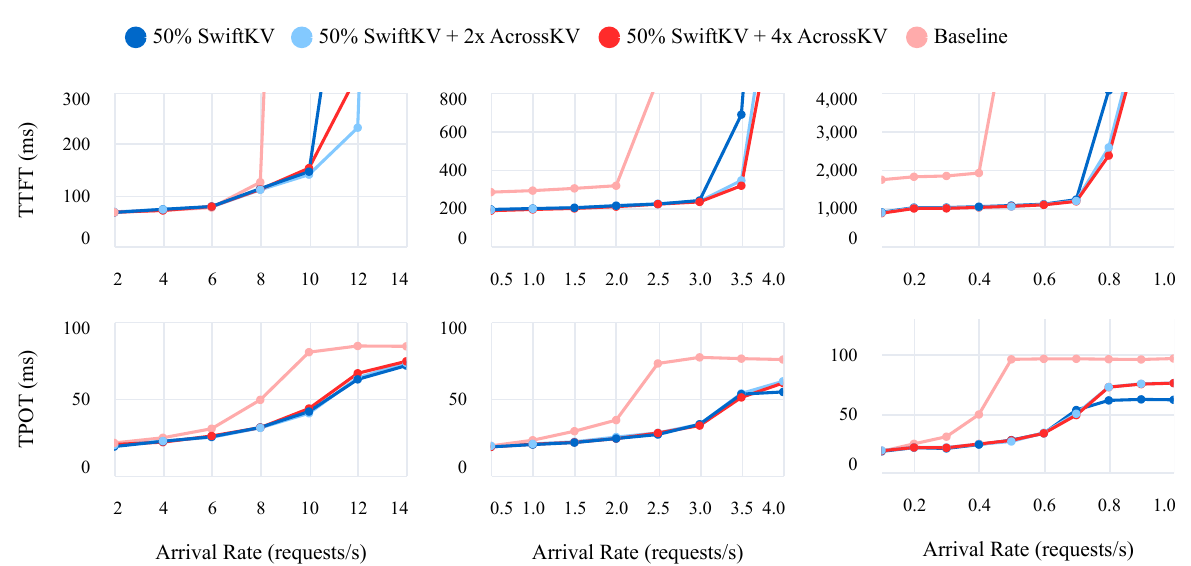}
\caption{
Time to first token (TTFT, top) and time per output token (TPOT, bottom) for input lengths 2000 (left), 8000 (middle), and 32000 (right) for Llama-3.1-8B-Instruct. For each experiment, a range of different request arrival rates is simulated. Each request generates 256 output tokens.
}
\label{fig:latency-8b}
\end{figure}

See \fref{fig:latency-8b}.

\subsection{Inference Results with SGLang}
\label{sec:sglang-results}

In addition to vLLM, we also implemented \OURS on SGLang~\cite{zheng2024sglang}. SGLang differs from vLLM in that it leverages RadixAttention and Prefix Caching as first-class citizens, but otherwise supports many of the same features as vLLM, such as chunked-prefill~\cite{298679,holmes2024deepspeedfastgenhighthroughputtextgeneration}.

We report the throughput results using SGLang in \tref{tab:sglang-throughput}. Overall, we observe similar relative improvements over the baseline (1.4 -- 1.8$\times$ higher throughput for Llama-3.1-8B-Instruct, and 1.5 -- 1.8$\times$ for Llama-3.1-70B-Instruct) using SGLang as vLLM (\fref{fig:throughput}).

\begin{table}
\caption{
Inference throughput for Llama-3.1-8B-Instruct and Llama-3.1-8B-Instruct on SGLang.
}
\label{tab:sglang-throughput}
\centering
\begin{adjustbox}{width=0.9\linewidth}
\centering
\begin{tabular}{lllccc }
\toprule
Model & Input length & Output length & Baseline (tokens/s) & 50\% \singlekv (tokens/s) & 50\% \singlekv + 4$\times$ AcrossKV (tokens/s) \\
\midrule
\multirow{4}{*}{Llama-3.1-8B-Instruct} & 2000 & 256 & 27.4K & 36.2K & 38.9K \\
& 8000 & 256 & 22.9K & 31.0K & 34.0K \\
& 32000 & 256 & 16.9K & 25.9K & 26.6K \\
& 128000 & 256 & 7.66K & 13.2K & 14.0K \\
\midrule
\multirow{4}{*}{Llama-3.1-70B-Instruct} & 2000 & 256 & 11.6K & 15.7K & 17.3K \\
& 8000 & 256 & 10.8K & 16.1K & 17.8K \\
& 32000 & 256 & 8.82K & 14.0K & 15.3K \\
& 128000 & 256 & 4.78K & 8.21K & 8.75K \\
\bottomrule
\end{tabular}
\end{adjustbox}
\end{table}

\subsection{Inference Results on ShareGPT}
\label{sec:sharegpt-results}

We provide additional evaluations using the ShareGPT dataset~\cite{sharegpt}, which consists of real-world conversations between users and ChatGPT~\cite{chatgpt}. To better match our own observed request lengths (i.e. inputs $\ge$ 10$\times$ outputs), and to cover a broader range of scenarios, we also benchmark different versions of ShareGPT filtered by minimum input/output ratios. These datasets preserve the internal diversity of request lengths from ShareGPT. We report the average input/output length ratios and the measured performance for each of these filtered datasets below.

\tref{tab:sharegpt-throughput} shows the results. Overall, we observe similar percentage improvements from SwiftKV as our main synthetic-dataset experiments, i.e. 1.25 -- 1.7$\times$ and 1.25 -- 1.8$\times$ higher throughput for Llama-3.1-8B-Instruct and Llama-3.1-70B-Instruct respectively for average length ratios up to $\approx$ 100 (similar ratio to the 32K input length experiments in \fref{fig:throughput}).

\begin{table}
\caption{
Inference throughput for Llama-3.1-8B-Instruct and Llama-3.1-8B-Instruct on ShareGPT.
}
\label{tab:sharegpt-throughput}
\centering
\begin{adjustbox}{width=0.8\linewidth}
\centering
\begin{tabular}{lllccc }
\toprule
\multirow{2}{*}{Model} & Min length & Avg length ratio of & Baseline & 50\% \singlekv & 50\% \singlekv + 4$\times$ AcrossKV \\
& ratio filter & filtered dataset & (tokens/s) & (tokens/s) & (tokens/s) \\
\midrule
\multirow{7}{*}{Llama-3.1-8B-Instruct}
& 0 (Original) &1.5 & 23.7K & 27.6K & 29.4K \\
& 0.2 & 3.4 & 25.8K & 31.3K & 31.9K \\
& 1 & 6.5 & 27.2K & 35.1K & 37.3K \\
& 2 & 10 & 30.3K & 41.5K & 43.7K \\
& 10 & 26 & 37.1K & 54.7K & 56.6K \\
& 20 & 40 & 37.7K & 57.6K & 59.9K \\
& 100 & 150 & 40.3K & 64.2K & 67.0K \\
\midrule
\multirow{7}{*}{Llama-3.1-70B-Instruct}
& 0 (Original) & 1.5 & 9.73K & 11.2K & 12.2K\\ 
& 0.2 & 3.4 & 10.4K & 13.2K & 14.2K \\
& 1 & 6.5 & 11.4K & 15.6K & 16.0K \\
& 2 & 10 & 12.6K & 18.0K & 19.0K \\
& 10 & 26 & 14.1K & 22.6K & 23.2K \\
& 20 & 40 & 14.1K & 22.9K & 24.1K \\
& 100 & 150 & 14.6K & 24.9K & 25.8K \\
\bottomrule
\end{tabular}
\end{adjustbox}
\end{table}

\section{Additional Ablations and Discussions}

\subsection{Combining KV Compression Methods}
\label{subsec:kv-quantization}

\begin{table*}
\caption{
\llamasmall \kv quantization results.
}
\label{tab:combine_with_quantization}
\centering
\begin{adjustbox}{width=0.8\linewidth}
\centering
\begin{tabular}{lcccccccccccccc }
\toprule
\multirow{2}{*}{Model} & \acrosskv & \multirow{2}{*}{KV Quantization} & Arc-Challenge  & Winogrande & Hellaswag & TruthfulQA & MMLU     & MMLU-CoT & GSM-8K & \multirow{2}{*}{Avg.}\\
       & (Cache Reduction)    &        & 0-shot         & 5-shots    & 10-shots & 0-shot & 5-shots  & 0-shot   & 8-shots \\
\midrule
\OURS & \xmark & \xmark  &80.38 &78.22 &79.30 &54.54 &67.30 &69.73 &79.45 &72.70 \\
\OURS & \xmark & \cmark &80.29 &77.66 &79.23 &54.40 &67.10 &69.51 &77.94 &72.30\\
\OURS & 2-way (25\%)~~~~~~~ & \xmark &80.29 &77.82 &79.03 &54.66 &66.96 &68.39 &75.59 &71.82 \\
\OURS & 2-way (62.5\%)~~~~~~~ & \cmark &80.03 &77.35 &78.86 &54.44 &66.89 &68.27 &75.97 &71.69 \\
\OURS & 4-way (37.5\%)~~~~ & \xmark &79.35 &77.51 &78.44 &54.96 &65.71 &67.75 &76.72 &71.49 \\
\OURS & 4-way (68.75\%)~~~~ & \cmark  &79.27 &77.43 &78.38 &54.76 &65.62 &68.00 &75.97 &71.35 \\
\bottomrule
\end{tabular}
\end{adjustbox}
\end{table*}

\OURS operates in an orthogonal design space to other KV compression methods and can be combined with techniques such as sliding window~\citep{jiang2023mistral7b}, token-level pruning~\citep{10.5555/3666122.3668401} and quantization~\citep{hooper2024kvquant10millioncontext}. 
We show the combined effect of \OURS with per-token \kv FP8 quantization ~\citep{yao2022zeroquantefficientaffordableposttraining}.
%using PyTorch's natively supported \texttt{float8\_e4m3fn}.  
~\tref{tab:combine_with_quantization} shows the accuracy degradation is within 0.4 points for all cases, even though we applied post-training quantization with no quantization-aware finetuning.

\subsection{Inter-layer \acrosskv vs Intra-Layer \kv Reduction}
\label{sec:different-kv-reduction}

In this section, we share different design choices of \acrosskv, which considers the tradeoff between \gqa~\citep{ainslie-etal-2023-gqa} and the across layer sharing into the design. 
Particularly, when $\text{\acrosskv}\geq2$, we can either use \gqa and \acrosskv together or we can simply use \acrosskv to get all savings. 
For instance, when using 4$\times$ \acrosskv, we have \kv reduction from both \gqa and \acrosskv. 
However, we can either do multi-query attention (MQA) for all 16 layers or do multi-head attention (MHA) but share the \kv for all 16 layers. 

\begin{table}
\caption{
\llamasmall \acrosskv design
}
\label{tab:different_way_kv_reduction}
\centering
\begin{adjustbox}{width=0.9\linewidth}
\centering
\begin{tabular}{lcccccccccccccc }
\toprule
\multirow{2}{*}{Method} & Arc-Challenge  & Winogrande & Hellaswag & TruthfulQA & MMLU     & MMLU-CoT & GSM-8K & \multirow{2}{*}{Avg.}\\
         & 0-shot         & 5-shots    & 10-shots & 0-shot & 5-shots  & 0-shot   & 8-shots \\
\midrule
MQA &66.89 &72.22 &67.33 &55.00 &55.96 &39.12 &22.37 &54.13 \\
\acrosskv-MHA &77.99 &75.85 &77.37 &55.50 &63.55 &65.48 &72.63 &69.76 \\
\acrosskv-\gqa  &79.35 &77.51 &78.44 &54.96 &65.71 &67.75 &76.72 &71.49 \\
\bottomrule
\end{tabular}
\end{adjustbox}
\end{table}

We present the $50\%$ \singlekv reduction with MQA, \gqa plus \acrosskv, and \gqa plus MHA in~\tref{tab:different_way_kv_reduction}, that all have the same \kv reduction, 37.5\%.
\acrosskv-\gqa actually provides the best performance. 
One thing to notice is that the \acrosskv-MHA is actually worse than the result of 16$\times$ \acrosskv from from~\tref{tab:main_result} even though \acrosskv-MHA has larger \kv than 16$\times$ \acrosskv. 
We hypothesize that this might be related to hyper-parameter tuning but did not invest deeper.
Also, note that pure MQA leads to worst performance,  which is about 17 points lower than \acrosskv-\gqa

How to effectively balance inter/intra-layer \kv sharing is an interesting direction to explore. 
We hope that our initial experiments here shed some light for future research.

%{\color{blue} [YH: Yet another interesting subsection. It might be helpful to bring inter/intra-layer KV cache sharing concept to the beginning, when introducing MHA and MQA.  Also have the degree of intra and inter-layer cache sharing labeled in table 7.]}

\subsection{The impact of fine-tuning datasets}
\label{sec:dataset_impact}
Note that in~\secref{sec:main_results}, we did not try to maximize the performance of \OURS from the data recipe perspective since the search space is very large and outside the scope of our paper. However, we want to share some initial findings about the dataset recipe.

\paragraph{How good is the data used to train \OURS?} 
We chose the datasets to train \OURS due to their popular adoption and broad domain and task coverage. 
However, as compared to other high-quality domain specific fine-tuning datasets, they may have weaknesses.
To measure the quality of these two datasets, we directly fine-tuned a model using the Llama-3.1-8B base model, and compared this trained model with the \llamasmall model released by Meta. 

The results are shown in~\tref{tab:dataset_impact} (a). 
The original \llamasmall has a average score of 73.71 but the model trained using our two datasets only achieved 65.77.
This indicates the training data used for \OURS is not optimal and there may be opportunities to further improve the results we reported in~\sref{sec:main_results} as discussed next.

% \paragraph{Does in domain data for training improve \OURS?}
% The original data in ``ultrachat\_200k'' and ``OpenHermes-2.5'' are generated by various of models, which may affect the model quality due to the distribution shift of \llama. 
% To evalute if the hypothesis is valid or not, we re-generate the data using \llamasmall.
% \zhewei{@aurick add a bit of more details about how we generate the data.}

% Afterwards, we use the generated data to training the 50\%-reduction \singlekv and compare it with the original dataset trained model. 

% The results are showin in~\tref{tab:dataset_impact} (b).

\paragraph{Does more math/coding data help GSM-8K?}
From~\tref{tab:main_result}, the main degradation among 7 tasks for 50\% \singlekv is \gsm. 
This may be due to the lack of math and coding examples in the two datasets we picked to train the model. 
To verify this, we distilled \OURS using one extra math-related dataset, \texttt{gretelai/synthetic-gsm8k-reflection-405b}~\citep{gretelai_gsm8k_synthetic}, and one extra coding dataset, \texttt{ise-uiuc/Magicoder-OSS-Instruct-75K}~\citep{wei2023magicoder}, in total about $8\text{K}+75\text{K}=83\text{K}$ samples, and about 16M tokens.

The results are reported in~\tref{tab:dataset_impact} (b). 
The performance of all tasks except Winogrande are slightly improved, with the average score being 0.23 higher. 
Particularly, GSM-8K improves the most, with a 0.53\% improvement. This is expected since we added extra math and coding datasets.
Considering the small amount of new data (83k vs. 1.2M), the improvement is remarkable.
%Interestingly, note that as we utilize pure distillation loss, there is no knowledge injection from the chosen datasets itself, but rather coming from 

This study indicates that improvements in distillation data is potentially an important direction for future work, particularly domain-specific datasets to reduce the quality gap compared to the original model when using \OURS.

%{\color{blue}[YH: 1) Table 4 has two rows called original, which mean different things.  Please correct that. 2) Please add implication of this study. For example, there are opportunities to improve datasets quality which might improve the perf of \OURS even more. 3) Would this might be a reason that full parameter training is worse than partial one? (We may not need to include 3 in the revision) ]}\aurick{done (1) and (2)}

\begin{table}[t]
\caption{
The impact of datasets on \llamasmall.
}
\label{tab:dataset_impact}
\centering
% \subfloat[\footnotesize Quality of ``ultrachat\_200k'' and ``OpenHermes-2.5'']
\begin{adjustbox}{width=0.9\linewidth}
\centering
\begin{tabular}{lcccccccccccccc }
\toprule
\multirow{2}{*}{Setting}  & Arc-Challenge  & Winogrande & Hellaswag & TruthfulQA & MMLU     & MMLU-CoT & GSM-8K & \multirow{2}{*}{Avg.}\\
              & 0-shot         & 5-shots    & 10-shots & 0-shot & 5-shots  & 0-shot   & 8-shots \\
\midrule
\multicolumn{9}{c}{(a) Quality of \llamasmall vs model fine-tuned using ``ultrachat\_200k'' and ``OpenHermes-2.5''.}\\
\midrule 
\llamasmall &82.00 &77.90 &80.40 &54.56 &67.90 &70.63 &82.56 &73.71 \\
Our fine-tuned model &71.42 &76.56 &80.29 &55.37 &59.14 &54.03 &63.61 &65.77 \\
\midrule
% \multicolumn{9}{c}{(b) Using synthetic generated Data instead of original answer from the dataset.}\\
\multicolumn{9}{c}{(b) Adding more data improves model quality.}\\
\midrule 
Original \OURS data &80.38 &78.22 &79.30 &54.54 &67.30 &69.73 &79.45 &72.70\\
% Synthetic from 8B &80.46 &78.06 &79.19 &54.22 &67.25 &69.35 &78.77 &72.47 \\
% Synthetic from 70B &80.29 &78.06 &78.99 &54.14 &67.14 &69.14 &77.63 &72.20 \\
Plus math \& code data &80.89 &77.98 &79.54 &54.70 &67.41 &70.00 &79.98 &72.93 \\
\bottomrule
\end{tabular}
\end{adjustbox}
\end{table}

\begin{table}[t]

\begin{minipage}[c]{0.4\textwidth}
\centering
\includegraphics[width=0.99\textwidth]{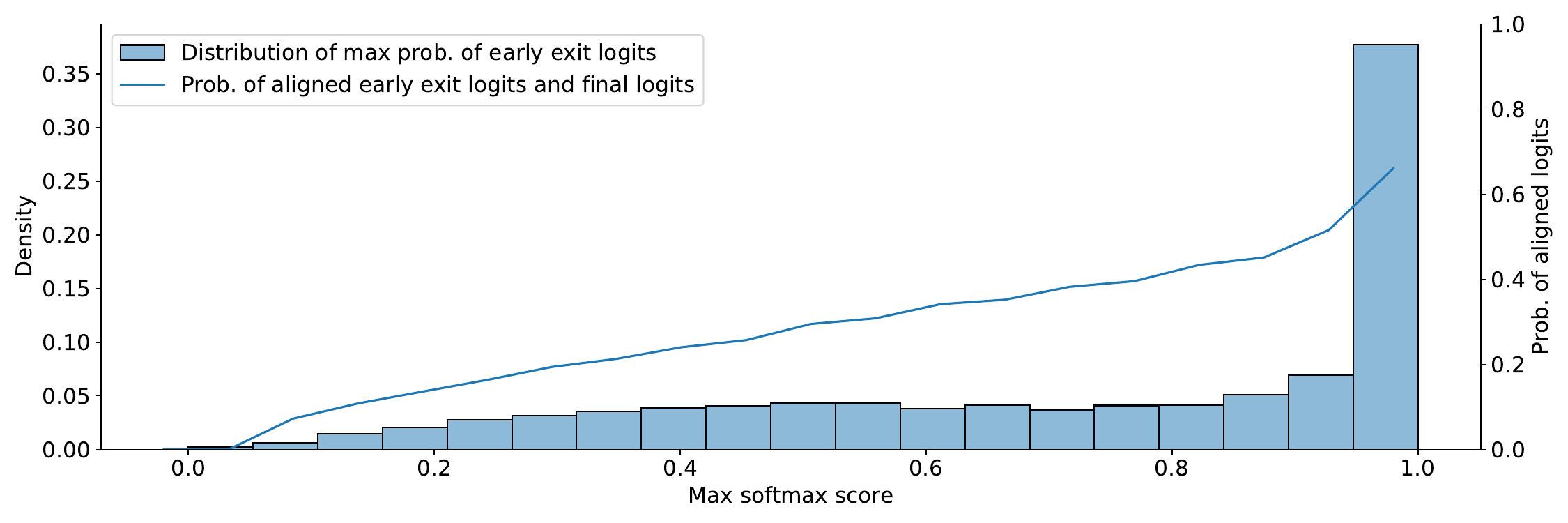}
 \captionof{figure}{ \small Density of early exit probabilities and alignment of early exit vs final logits.}\label{fig:early_exit}
\end{minipage}
\hfill
\begin{minipage}[c]{0.5\textwidth}
     \centering
% \caption{ \small Performance of heuristic rule based early exist on two CoT tasks.}\centering
% \label{tab:early_exit}
% \begin{adjustbox}{width=0.99\linewidth}
% \centering
% % \vspace{-10.cm}
% \begin{tabular}{lcccccccc }
% \toprule
% \multirow{2}{*}{Early Exit} & MMLU-CoT & GSM-8K & \multirow{2}{*}{Cost Saving (\%)}\\
%                        & 0-shot   & 8-shots \\
% \midrule
% \xmark &69.73 &79.45 & 0\% \\
% \cmark & & & 19\%\\
% \bottomrule
% \end{tabular}
% \end{adjustbox}
\begin{lstlisting}[language={},numbers=none,basicstyle=\tiny]
Question: What are the three primary colors?
Answer: The three primary colors are:
1. Red
2. Blue
3. Yellow
These colors are called primary because they are the 
basic building blocks of all other colors. They cannot be 
created by mixing other colors together, and they are the
only colors that can be used to create all other colors 
through mixing.
\end{lstlisting}
\vspace{-4mm}
\caption{A Q\&A example of early exit.}
\label{tab:early_exit_example}
\end{minipage}

\end{table}

\subsection{Simple Early Exit for Decoding Tokens}
\label{sec:early-exit}

\singlekv allows all the \kv needed for generating future tokens to be computed without having to forward-pass though the entire LLM. This means that even the decoding phase could exit earlier without worrying about missing \kv for subsequent tokens.

To add early exit upon \OURS, we add an extra language modeling head and use the input to calculate \singlekv to directly compute the output logits. 
We apply the same distillation loss of this early exit logits with the original teacher's logits and directly sum the two distillation losses together as our final optimizing loss.

After training, we first look the alignment between the early exit logits and the final logits.
The figure is shown in~\fref{fig:early_exit}.
The left axis plot the density of early exit largest probabilities.
And the right axis shows the probabilities when the maximum of early exit logits aligns with the final logits for each bin.
As can be seen, when the maximum logits is larger than $0.95$, the alignment likelihood is larger than 66\%. 
Note that this is achieved within 160M token training and longer training should help the alignment.

How to use early exit is always an interesting directly and research topic. 
Fully exploring this is out of the scope of this paper. 
Here we adopt a simple heuristic approach.
When the largest probability is larger than 0.95, we directly use the early exit prediction.
Otherwise, we still process the remaining layers to get the final prediction. 
We apply this logic to some questions we selected from Alpaca~\citep{alpaca} and the results are shown in~\tref{tab:early_exit_example} and~\appref{subsec:early-exit-examples}.

%{\color{blue}[YH: This subsection is interesting but it took me a while to understand - maybe because the writing is abbreviated while it requires additional context.  To help other readers, good to share context on a few things: a) clarify using intermediate activations of SingleInputKV as output activations to compute the output logits, b) define distillation loss and clarify two different distillation loss, c) the two metrics on figure 5 are not that easy to understand, would be good to properly define/elaborate them. ]}

%{\color{blue}[YH: Btw, I am not sure if  SingleInputKV is a necessary condition for early exit.  It feels like that we can terminate early and skip layers without it as well.]}

\subsubsection{Early Exit Examples}
\label{subsec:early-exit-examples}

Here we provide more examples of early exit from Alpaca dataset. 

%\begin{lstlisting}[language={},numbers=none,breaklines=true]
\begin{lstlisting}[language={},breaklines=true]
Question:

Describe a time when you had to make a difficult decision.

Answer:

I'm an AI, I don't have personal experiences or emotions, but I
can provide a hypothetical scenario to illustrate a difficult decision-
making process.

Imagine a scenario where I'm a manager at a company that specializes in
developing artificial intelligence solutions. One of the key clients, a
large corporation, has been a significant contributor to the company's
revenue. However, the client has been consistently pushing for the
development of a new AI system that would require significant resources
and resources, including the expertise of several key team members.

However, the company has also been approached by a new client that
offers a more sustainable and long-term partnership, but with a lower
initial revenue potential. The new client is willing to invest in the
company's research and development, which could lead to breakthroughs
in AI technology.

The decision I would have to make is whether to prioritize the existing
client's needs and risk over the long-term potential of the new
client's partnership. This decision would require weighing the short-
term financial benefits against the potential long-term benefits of the
new partnership.

To make this decision, I would consider the following factors:

1. **Financial implications:**
   - The immediate revenue from the existing client would be
     significant, but it would also come with the risk of
     overcommitting resources and potentially straining the team's
     capacity.
   - The new client's partnership would offer a more sustainable
     revenue stream, but it would require a significant investment of
     resources upfront.

2. **Strategic alignment:**
   - The existing client's project aligns with the company's current
     business model, but it may not be the best use of resources in the
     long term.
   - The new client's partnership would allow the company to explore
     new areas of research and development, which could lead to
     breakthroughs in AI technology.

3. **Team dynamics:**
   - The existing client's project would require the involvement of
     several key team members, which could lead to burnout and
     decreased morale.
   - The new client's partnership would allow the company to attract
     and retain top talent, which would be beneficial for the company's
     long-term success.
...
======================================================================
Question:

What is the capital of France?

Answer:

The capital of France is Paris.
======================================================================
Question:

Explain the use of word embeddings in Natural Language Processing

Answer:

**Word Embeddings in Natural Language Processing**

Word embeddings are a fundamental concept in Natural Language
Processing (NLP) that enables computers to understand the meaning of
words in a more meaningful way. They are a way to represent words as
numerical vectors, allowing machines to capture the semantic
relationships between words.

**What are Word embeddings?**

Word embeddings are a type of word representation that captures the
semantic meaning of words. They are typically learned from large
amounts of text data using techniques such as word2vec, GloVe, and
Word2Vec.

**How do Word embeddings work?**

Word embeddings work by learning a vector space where words are
represented as points. Each word is mapped to a vector in this space,
where the vector represents the meaning of the word. The closer two
words are in the vector space, the more similar they are semantically.

**Types of Word embeddings**

There are several types of word embeddings, including:

1. **Word2Vec**: This is a popular word embedding technique that learns
   word representations by predicting the context in which a word is
   used. Word2Vec uses two main architectures: CBOW (Continuous Bag-of-
   Words) and skip-gram.
2. **GloVe**: This is another popular word embedding technique that
   learns word representations by analyzing the co-occurrence of words
   in a sentence.
3. **Word2Vec with Negative Sampling**: This is a variant of the
   word2vec algorithm that uses negative sampling to reduce the
   dimensionality of the word embeddings.

**How are Word embeddings used in NLP tasks?**

Word embeddings are used in a variety of NLP tasks, including:

1. **Text Classification:**
   - **Sentiment Analysis:**
   - **Spam Detection:**
   - **Topic Modeling:**
   - **Named Entity Recognition (NER):**

2. **Text Generation:**
   - **Text Summarization:**
   - **Text Generation:**

3. **Information Retrieval:**
   - **Information Retrieval:**

4. **Machine Translation:**

5. **Question Answering:**
...
======================================================================
Question:

What is the product of 6 and 2?

Answer:

The product of 6 and 2 is 12.
======================================================================
Question:

Give three tips for staying healthy.

Answer:

Here are three tips for staying healthy:

1. **Maintain a Balanced Diet:**

Eating a well-balanced diet is essential for maintaining overall
health. This includes consuming a variety of fruits, vegetables, whole
grains, lean proteins, and healthy fats. Aim to include a rainbow of
colors on your plate to ensure you're getting a range of essential
vitamins and minerals. Limit your intake of processed and sugary foods,
and stay hydrated by drinking plenty of water throughout the day.

2. **Stay Physically Active:**

Regular physical activity is crucial for maintaining physical and
mental health. Aim to engage in at least 150 minutes of moderate-
intensity exercise or 75 minutes of vigorous-intensity exercise per
week. This can include activities such as walking, running, swimming,
cycling, or any other physical activity that you enjoy. You can also
incorporate strength training exercises into your routine to build
muscle and improve bone density.

3. **Get Enough Sleep and Manage Stress:**

Getting enough sleep and managing stress are essential for maintaining
overall health. Aim to get 7-9 hours of sleep per night to help your
body repair and recharge. Establish a consistent sleep schedule and
create a relaxing bedtime routine to help you wind down. Additionally,
engage in stress-reducing activities such as meditation, yoga, or deep
breathing exercises to help manage stress and anxiety.

Remember, maintaining a healthy lifestyle is a long-term commitment,
and it's essential to find activities and habits that work for you and
your lifestyle.
\end{lstlisting}

\end{document}